# Cyberbullying Detection for Low-resource Languages and Dialects: Review of the State of the Art

Tanjim Mahmud[a,b,*], Michal Ptaszynski[a,*], Juuso Eronen[a] and Fumito Masui[a]

[a]*Text Information Processing Lab, Kitami Institute of Technology, Kitami, Japan*
[b]*Dept. of Computer Science and Engineering, Rangamati Science and Technology University, Bangladesh*

ABSTRACT

The struggle of social media platforms to moderate content in a timely manner, encourages users to abuse such platforms to spread vulgar or abusive language, which, when performed repeatedly becomes **cyberbullying** – a social problem taking place in virtual environments, yet with real-world consequences, such as depression, withdrawal, or even suicide attempts of its victims. Systems for the automatic detection and mitigation of cyberbullying have been developed but, unfortunately, the vast majority of them are for the English language, with only a handful available for **low-resource languages**. To estimate the present state of research and recognize the needs for further development, in this paper we present a comprehensive systematic survey of studies done so far for *automatic cyberbullying detection in low-resource languages*. We analyzed all studies on this topic that were available. We investigated more than seventy published studies on automatic detection of cyberbullying or related language in low-resource languages and dialects that were published between around 2017 and January 2023. There are 23 low-resource languages and dialects covered by this paper, including Bangla, Hindi, Dravidian languages and others. In the survey, we identify some of the research gaps of previous studies, which include the lack of reliable definitions of cyberbullying and its relevant subcategories, biases in the acquisition, and annotation of data. Based on recognizing those research gaps, we provide some suggestions for improving the general research conduct in cyberbullying detection, with a primary focus on low-resource languages. Based on those proposed suggestions, we collect and release a cyberbullying dataset in the Chittagonian dialect of Bangla and propose a number of initial ML solutions trained on that dataset. In addition, pre-trained transformer-based the BanglaBERT model was also attempted. We conclude with additional discussions on ethical issues regarding such studies, highlight how our survey improves on similar surveys done in the past, and discuss the usefulness of recently popular AI-enhanced tools for streamlining such scientific surveys.

## 1. Introduction

Since the beginning of 21st century Social Networking Services (SNS), such as Facebook or Twitter have started to gain in popularity. With the burst of popularity of SNS especially in the second decade of the 2000s, a large part of everyday human communication has been taking place on the Internet[1,2,3]. Some of the most extensively utilized social networking services include Facebook, YouTube, WhatsApp, Instagram, Twitter or Reddit[4]. With nearly three billion monthly users, Facebook can be perceived as one of the most impressive inventions of Information and Communication Technology (ICT), connecting numerous people online, facilitating sharing of information related to one's personal life, uploading photos and videos, writing messages, viewing videos, or sharing links to recent news pieces or other information. With this, the users share their opinions, emotions, and feelings in their own respective languages. Consequently, each year the volume of Facebook, Twitter, and other SNS users has been increasing (Braghieri et al., 2022). However, as much as this gain in SNS popularity has had its merits, such as allowing more people to frequently connect through the Internet, it also has had some crucial demerits, with some studies claiming that SNS such as

[1]https://ourworldindata.org/grapher/ict-adoption
[2]https://ourworldindata.org/grapher/users-by-social-media-platform
[3]https://ourworldindata.org/grapher/daily-hours-spent-with-digital-media-per-adult-user
[4]https://www.statista.com/statistics/272014/global-social-networks-ranked-by-number-of-users/

Facebook actually directly caused a global decrease in the mental health of young people (Ptaszynski and Masui, 2018; Urbaniak et al., 2022a). One of the reasons for this decrease has been cyberbullying, a malicious activity where users with ill intent use abusive language in their posts or comments to insult and humiliate other users (Urbaniak et al., 2022a). The prevalence of cyberbullying in various countries is increasing, with countries like India, or Brazil struggling the most with the problem[5,6]. Because of that, the awareness of the problem of cyberbullying has also greatly increased in the last decade[7]. The various social media platforms have started attempts to mitigate this global problem by either allowing users to report potentially harmful content or trying to detect such unwanted content with the help of moderators or automatically.

Protecting everyday Internet users, especially young users, from cyberbullying has thus become an important matter. With the global increase of the problem of cyberbullying, it has been pointed out that the use of technology has become indispensable if any realistic mitigation of the problem is to be achieved (Ptaszynski and Masui, 2018). One of the popular means to automate the detection of cyberbullying has been by using Machine Learning (ML) and Deep Learning (DL) algorithms which allow for the detection, or correct classification of new relevant instances after training them with a sufficient amount of data (Eronen et al., 2021, 2022; Mahmud et al., 2023).

Especially with regards to the automatic detection of cyberbullying, recently there has also been more and more available data to develop such automatic cyberbullying detection systems[8]. However, most of the data is available for high-resource languages, such as English, or German, with only a small portion of data being available for low-resource languages, such as Bengali, Hindi, or Urdu.

Some studies have tried to overcome this issue by, e.g., translating data from high-resource languages, like English, to low-resource like Bengali (Jahan et al., 2019; Islam et al., 2021). However, such approaches are problematic due to the errors caused by machine translation systems used in such approaches. Therefore a more robust and precise approach is needed to help detect cyberbullying, especially in low-resource languages.

To lay the groundwork for proposing such an approach, we present our survey in research done in automatic cyberbullying detection, with a specific focus on research done in low-resource languages. We analyzed a substantial amount of previous studies to find out what are the problems facing such research, e.g., how difficult it is to collect data for such languages, which classification methods have been used so far, which ones have worked the best, etc. From this analysis, we drew conclusions on the most promising approaches toward automatic cyberbullying detection in low-resource languages.

## 1.1. Research objectives and contributions

Below is the list of all the research objectives.

1. To provide a full view of current studies of the methods used to detect cyberbullying or related topics on social media, with a specific focus on research done for low-research languages.
2. To find the most prevalent research gaps and provide a proposal of good practices in such research.
3. Based on the good practices, to develop criteria for data annotation with cyberbullying-related categories.
4. Based on the developed criteria, to develop an initial dataset for detecting cyberbullying on social media in the Chittagonian dialect of Bangla – one of the underrepresented languages on social media.

Below are key contributions of this paper.

1. We explored more than seventy existing studies on cyberbullying or related topics detection in low-resource languages and dialects from 2017 to 2023.
2. We identified the most common research gaps in such research and provided a list of good practices for future efforts.
3. We created the first dataset of cyberbullying social media entries in the Chittagonian dialect of Bangla and manually annotated it with highly trained annotators.
4. We performed a series of experiments to develop a number of strong baselines for the classification based on this dataset.
5. We performed an extensive examination on how to perform similar survey studies in the future, including a discussion on the applicability of Artificial Intelligence-based tools for survey writing support.

---

[5]https://www.comparitech.com/internet-providers/cyberbullying-statistics/
[6]https://cyberbullying.org/cyberbullying-statistics-age-gender-sexual-orientation-race
[7]https://www.statista.com/statistics/293192/cyber-bullying-awareness-in-select-countries-worldwide/
[8]https://hatespeechdata.com

**Table 1**
Distinctions of various manifestations with cyberbullying

| Manifestations | Characteristics of Manifestations | Distinction from cyberbullying |
|---|---|---|
| Hate Speech | The target of hate speech is to spread mistrust, hostility, or prejudice. | People of any age, gender, or ethnicity can be the targets of cyberbullying. It can happen on a variety of online platforms, such as social media, messaging services, forums, and email. |
| Abusive Language | The goal of abusive language is to hurt the target's feelings. | Intentional and persistent acts of aggression against particular people are considered cyberbullying. |
| Offensive Language | The expression of strong opinions or feelings targeted at a particular individual or group is the main characteristic of offensive language | Inappropriate or disrespectful language can be used without necessarily being directed at a specific person. This is referred to as offensive language. |

This paper is assembled as follows. In section 2 we present the brief description of the problem of cyberbullying, its background, and its types and current psychological studies. In section 3 we specifically summarize machine learning methods used in low-resource language-based automatic cyberbullying detection as well as research closely related to this field such as abusive language detection, hate speech language detection and offensive language detection, specifically done in low-resource languages. In section 4 we present the initial attempt at creating and testing a dataset for cyberbullying detection in the Chittagonian dialect of Bangla language In section 5 we disclose ethical considerations during the process of data collection as well as discussion about artificial intelligence-based support for writing scientific field surveys. Finally, in section 6 we conclude the paper and outline future directions.

## 2. Research on cyberbullying in countries of low-resource languages

### 2.1. Cyberbullying – overview of the problem

Cyberbullying, or bullying on the Internet, is defined as sending threatening or humiliating comments to another user through the use of Information Technology, such as computers, tablets, or mobile phones (Willard, 2007a). It is spread through social media, online forums, or chat systems such as those in online games where users can read, interact with, or exchange content (Patchin and Hinduja, 2006). Specifically, cyberbullying is the act of sending, posting, or circulating negative, hurtful, or malicious content about another user (Maher, 2008). It might also entail revealing the user's sensitive or private information in a way that makes them feel ashamed or embarrassed. Sometimes cyberbullying can develop into direct criminal acts, such as death threats (Bauman et al., 2013).

In general, cyberbullying is defined as the practice of harassing or verbally assaulting a private person, or a small group of private persons repeatedly online while using offensive, hateful, or abusive language (Ptaszynski and Masui, 2018). Thus, to detect cyberbullying, it is necessary to be able to spot offensive, abusive, and hateful language. However, there are some distinctions of those three kinds of language with cyberbullying behavior (See Table 1), such as speaking negatively about a group of people or a person with regards to their affiliation to the group characterized by their race, ethnicity, religion, gender, sexual orientation, or other characteristics which is considered hate speech[9]. Threats, racial slurs, and statements that dehumanize others are all examples of hate speech[10]. Abusive language is any form of communication that singles out, disparages, or insults another person. This includes verbal abuse, insults, and name-calling[11]. Offensive language is any statement that might offend or disturb another person. This might include obscene language, offensive claims, or profanity[12]. This way both abusive and offensive language are often tools used by cyberbullies. Hate speech on the other hand, differs from cyberbullying by aiming at a larger group because of their general characteristics.

---

[9]https://hatebusters.erasmus.site/hatespeech-and-cyberbullying/
[10]https://www.public.io/case-study/online-safety-data-initiative
[11]https://www.lawinsider.com/dictionary/abusive-language
[12]https://www.lawinsider.com/dictionary/offensive-language

Glasser, who developed the choice theory in 1996 (Glasser, 1997a), asserts that children are accountable for their own behavior and frequently make sensible judgments when their fundamental needs are met. Students occasionally make poor choices that might result in inappropriate behavior. The choice theory puts in its foundations the need for freedom, amusement, love, and belonging (Glasser, 1997b). Glasser conducted a study on issues that crop up in schools, which he used to create reality therapy and choice theory. This involved conducting research to stop cyberbullying actions that were supported by choice theory and reality therapy (Tanrikulu, 2014).

Sullo (1997) asserts that humans may not be able to notice their surroundings if their physiological needs are not supplied. Because children's fundamental needs must be satisfied in a learning environment for them to feel connected to the school environment, the choice theory is pertinent to bullying issues (Glasser, 1999). If these demands are not fulfilled, students may begin acting in an improper manner.

According to O'Brennan, Bradshaw, and Sawyer, bullying can result in social-emotional problems, and both the bully and the victim may feel afraid or alone in a school environment (O'Brennan et al., 2009).

According to Waasdorp and Bradshaw, since every child develops at a different rate, elementary school children who experience bullying may turn to their parents or teachers for support while middle school and high school students may turn to their peers for support or decide to handle it on their own (Waasdorp and Bradshaw, 2009). Therefore, school administrators should stop bullying by not allowing pupils to handle it on their own, but rather allow teachers and educators to use effective educational procedures (O'Brennan et al., 2009).

Because of its global presence, bullying on social media has the potential to be much more damaging than face-to-face bullying (Braghieri et al., 2022). Cyberbullying frequently occurs on various SNS, including Facebook, YouTube, WhatsApp, Twitter or Reddit, hindering the quality of SNS experience for many regular users (Urbaniak et al., 2022a).

Cyberbullying can lead to a number of negative outcomes, including impaired psychological (Kowalski et al., 2012; Nixon, 2014) and physical well-being (Edition et al., 2013; Sarris, 2018), poor self-esteem, hopelessness (Schodt et al., 2021) and anxiety (Hellfeldt et al., 2020), and even suicidal thoughts (Krešić Ćorić and Kaštelan, 2020) and suicide attempts (Samaneh et al., 2013). Hence, cyberbullying has evolved into an epidemic that requires immediate attention. According to Willard (2007a), there are numerous ways cyberbullying can take place on the Internet (Willard, 2007b; League, 2011), including the following:

a Commencement of an online fight.
b Regularly sending rude or nasty comments to specific users.
c Bullies disguising themselves as another person in order to carry out malicious objectives.
d Trolling, also known as baiting, which is a deliberate posting of words that support other commenters in an emotionally charged discussion in an effort to spark controversy, even while the remarks do not actually reflect the poster's opinion.
e Denigration, or spreading gossip and rumors in order to harm someone's reputation.

Additionally, with the proliferation of Unicode and the growing use of the Internet, low-resource languages and dialects have also been used more widely on social media, with cyberbullying also having a crippling effect on the population of users using such languages in their everyday life.

## 2.2. What are low-resource languages?

English, Chinese, Spanish, French, Japanese, and other languages of developed countries are examples of high-resource languages (Niu et al., 2022).

On the other hand, the term "low-resource language" refers to languages that have a small number of available resources, such as written literature, or other transcripts, as well as an Internet presence. This includes endangered languages, but also languages that are in general less represented on the Internet, or that have fewer resources available for their analysis (Akhter et al., 2018). Analysis of low-resource languages is currently one of the most difficult tasks in Natural Language Processing (NLP). It is difficult to analyze and manipulate such languages because they usually lack the requisite training qualities, such as labeled data or a large enough community of native speakers or experts, etc. (Nowakowski et al., 2019, 2023).

NLP tools underwent a significant change in the recent decade, as rule-based methods were replaced with statistical methods, which recently became replaced with data-hungry neural network-based approaches. The great majority of languages in the world are under-studied because just about 20% of the estimated 7000 languages spoken worldwide are the focus of the majority of current NLP research (Joshi et al., 2020). Although the term "low-resource language" has not been sufficiently precisely defined, it is regularly used to describe the languages of the above-mentioned characteristics. It can include those that are less privileged, less computerized, less frequently taught, and less studied, but they are not the only ones. Specifically, in the context of our study, we define languages referred to as low-resource languages as those for which there is not enough data to make high-performance statistical and neural network-based methods.

Moreover, focusing on low-resource languages in NLP research is significant for many reasons. Africa and India alone are home to more than 2.5 billion people, making up about 2000 low-resource languages. The creation of technologies for these languages is of high importance in the digital age, as it would significantly increase the range of available potential occupations. Moreover, making methods made for such languages accessible to a larger population would help such languages thrive and be preserved, and even improve emergency responses, which would allow such NLP methods to benefit the global population.

## 2.3. Cyberbullying in areas and countries of low-resource languages

The problem of cyberbullying has become so prevalent that it is estimated that anyone, anywhere, at any time could experience cyberbullying. For instance, in research by Beran and Li (2005) that involved 432 junior high school students from mid-level classes (classes 7 through 9) from nine junior high schools of ethnically diverse regions in Calgary, Canada, in 2004, more than two-thirds of students had heard about cyberbullying, and nearly a quarter of them had actually experienced it.

A related study based on 177 Canadian seventh-graders, discovered a comparable tendency (Beran and Li, 2007). Fifteen percent of the pupils specifically admitted to intimidating other classmates online. More than 40% of the cyber-victims had no idea who the bullies were, indicating that in this kind of behavior, perpetrators are often anonymous or at least unknown to the victims. Less than 35% of the witnesses also mentioned the incident to an adult.

The results of a global Ipsos survey [13], which included participants from 28 different countries, demonstrated that the number of parents aware that their children had experienced cyberbullying is increasing. Between March 23 and April 6, 2018, interviews were held with 20,793 adult respondents (ages 16 to 64 worldwide and 18 to 64 in the US and Canada).

Particularly noteworthy are the results for India [14], where the proportion of Indian parents who indicated they were aware their children had at least occasionally encountered cyberbullying greatly increased between 2011 and 2018. Additionally, it appears that more parents in Europe and the Americas are becoming aware of the negative effects that cyberbullying has had on their children or that they are experiencing more such online assaults[15].

In Bangladesh, women make up over 80% of the victims of cyberbullying between the ages of 14 and 22, while the majority of online abusers are teenagers between the ages of 16 and 17 [16].

Increased Internet and mobile device usage in India puts children at risk for cyberbullying (Patchin and Hinduja, 2006). In a study of 174 middle school children in Delhi, 17% of the participants reported having personally experienced cyberbullying, while 8% acknowledged perpetrating it. However, complaints of physical bullying, fighting, or victimization by either party were made in 16%, 12%, and 17% of the cases, respectively. Compared to women, men were more likely to employ physical force and intimidation. Women also had a higher chance of becoming the targets of aggressive behavior both locally and online[17].

As for another country of an under-represented language, the number of Internet users in Indonesia, increased by about 52% from 2016 to 2021[18]. Numerous studies have demonstrated that cyberbullying conduct has detrimental impacts on both the victim (Urbaniak et al., 2022a) as well as the bullies themselves (Zuckerman, 2016; Urbaniak et al., 2022b).

Over the past ten years, social media use has also increased in the Middle East. According to the most recent Internet statistics, 64.5 % of people in the Middle East use social networking sites, with 28% of them using Twitter (Donegan, 2012). With one hundred million people, or close to 50% of the population of the Middle East, being between the ages of 15 and 29, cyberbullying has gradually become a significant issue for the users in those ages also in the Arab world (Mouheb et al., 2019).

---

[13]https://www.ipsos.com/en/global-views-cyberbullying
[14]https://www.statista.com/statistics/293136/share-of-parent-awareness-of-their-child-being-cyber-bullied-worldwide-
[15]https://www.comparitech.com/internet-providers/cyberbullying-statistics/
[16]https://www.thedailystar.net/country/news/80-cyberbullying-victims-are-women-cyber-crime-division-dmp-2009017
[17]https://pubmed.ncbi.nlm.nih.gov/29061413/
[18]https://www.statista.com/statistics/254456/number-of-internet-users-in-indonesia/

## 3. Machine learning studies in low-resource language-based automatic cyberbullying detection

In recent years gains have been made in various NLP tasks, especially due to the development of the deep learning models (Dadvar and Eckert, 2020). A number of various machine learning and deep learning models were proposed for cyberbullying detection and related tasks (e.g., hate-speech detection, offensive language detection, etc.), including models based on several different types of word representations applied in classification (Ishmam and Sharmin, 2019; Eronen et al., 2021). Below, we describe a number of studies aimed at developing such models, especially for low-resource languages.

### 3.1. General overview of the field

Finding solutions for online harassment, e.g., by detecting abusive language on the Internet has become a popular research topic in recent years. Although a great deal of work has been done in English (Sazzed, 2021; Urbaniak et al., 2022a), and other languages with high representation on the Internet, such as Japanese (Ptaszynski and Masui, 2018; Ptaszynski et al., 2019; Eronen et al., 2021), Dutch, German (Eronen et al., 2022), etc., recently, an increasing number of works have involved other, much less represented languages, such as Bangla.

Since languages are diverse, and harassment can be expressed very differently even in languages from the same language family (see the comparison between Japanese and Korean in Eronen et al. (2022)), detecting cyberbullying, especially in low-resource languages is not an easy task. Each language has its own way of expressing heightened emotions, and it also depends on how people are using the words, what is the culture, and the behavior of people living in certain areas. Different types of linguistic and deep learning techniques are proving useful to mitigate the issue (Orabi et al., 2018). Nowadays, researchers are approaching this problem for many local and low-resource languages. A number of researchers worked in the Arabic language, which is widely spoken by Saudi Arabians and other people of the Middle East (Haidar et al., 2018; Mouheb et al., 2019; AlHarbi et al., 2019, 2020; Alsafari et al., 2020; Nayel, 2020; Alshalan and Al-Khalifa, 2020). This language was one of the first in which creation was initiated of offensive dialect datasets in the Arabic language, extracted from three social media platforms, namely, Facebook, Twitter, and YouTube. Researchers looked at such features as the lexical structure of sentences, or the use of emoji in abusive language samples. Alongside evaluating the performance of the classifiers in cross-platform and cross-domain environments, the researchers often used traditional machine learning algorithms like Support Vector Machines as classifiers (Samaneh et al., 2013; Eronen et al., 2021).

To fulfill the research objectives presented in section 1.1, we have discussed relevant academic works on the detection of cyberbullying in low-resource languages and dialects released between 2017 and 2023. In order to perform this research, we used the systematic review process (Moher et al., 2015; del Amo et al., 2018; Mengist et al., 2020). Specifically, the review process used in this study includes various steps as shown in Figure 1.

Using the systematic review approach, we selected 76 papers that were specifically on the topic of automatic detection of cyberbullying (including related phenomena, e.g., hate speech, toxicity, etc.) in low-resource languages, from a total number of 1,302 related publications written on harmful language detection in general (see Figure 2 for details).

The majority of authors published in venues related to IEEE (23.1%), Springer (14.1%), and ScienceDirect (11.5%),with the remaining minority scattered around such venues as ACM, MDPI and others (see Fig. 3 for details).

From the analysis, we found that an overall number of 258 authors contributed to the research in this area in this specified period. Additionally, we found out that the overall number of authors publishing in the area greatly increased in 2019 (see chart in Fig. 4). This could be influenced by the widespread of the Transformers architecture, especially multilingual models based on this architecture, such as mBERT, or XLM-Roberta.

However, the majority of authors published only one paper, as shown in Table 2. A few authors were associated with two (AlHarbi et al., 2019; Akhter et al., 2020; Saha et al., 2021; Dewani et al., 2021), or even four (Van Huynh et al., 2019) publications as shown in Figure 5. This shows that although there are many authors making initial attempts in the area, there is still lack of a longitudinal research done by the same research group for one specific low-resource language.

### 3.2. Studies on cyberbullying detection in various low-resource languages

In the past five years there have been a number of studies on automatic cyberbullying detection in many languages considered as low-resource, under-resourced, or under-represented on the Internet.

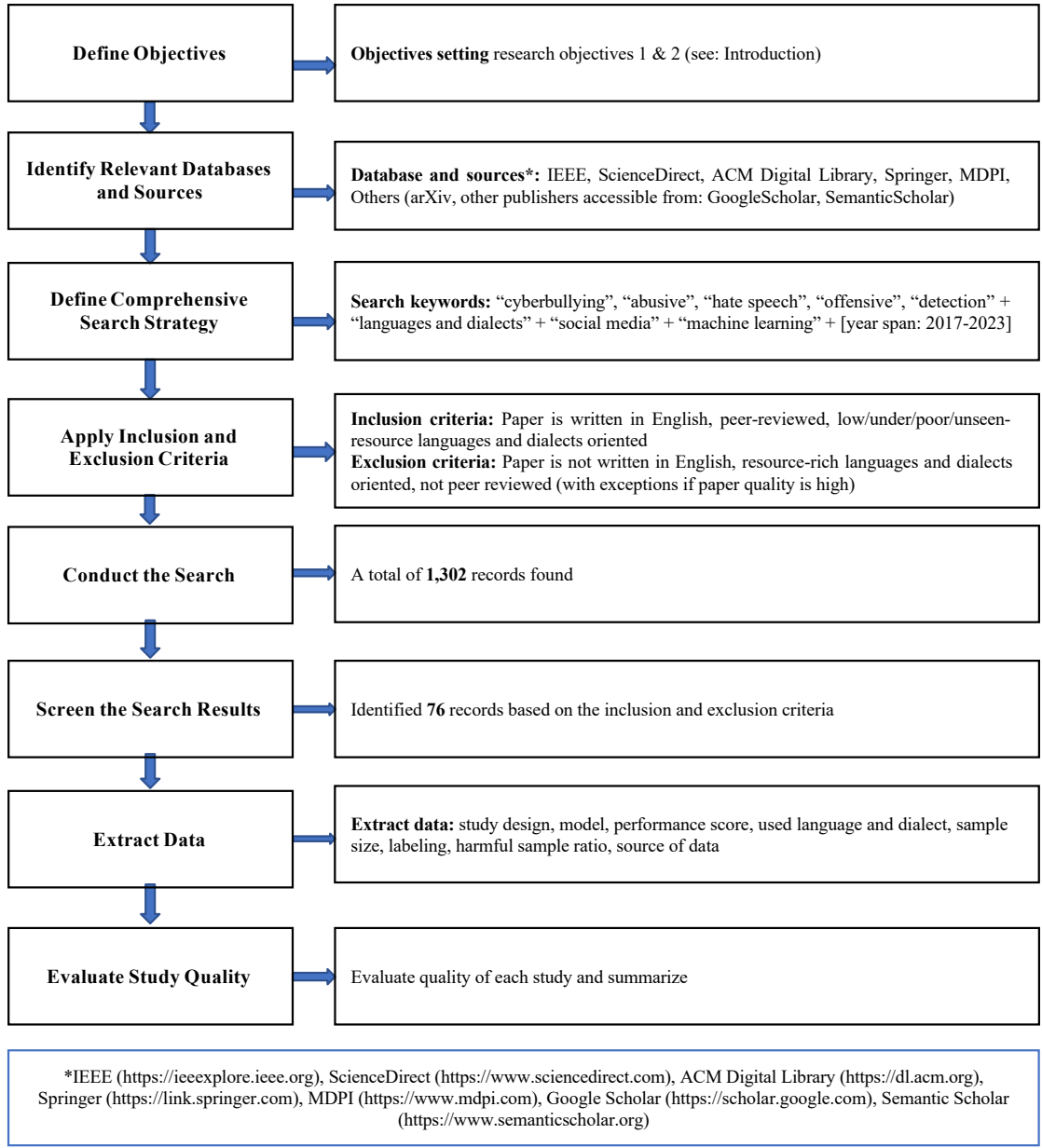

**Figure 1:** A rigorous research identification methodology.

Being the worlds fifth most spoken native language, Bangla/Bengali is used by more than 300 million users in their daily life[29]. Since it is the era of social networking, people are tending to use this language in their social media platforms like Facebook, where they can share any content including cyberbullying, harassment, and threats etc. Nevertheless, it is still regarded as a low-resource language. The lack of sufficient data for this language results in poor performance in various NLP tasks, such as text classification, but also machine translation, which is one of the main research problems in working with this language (Anand and Eswari, 2019). Marathi (Pandharipande, 2021) is an Indo-Aryan language spoken by 80 million people, the vast majority of whom are Indian. Marathi (Gaikwad et al., 2021), Tamil, Malayalam, and the Kannada language (Saha et al., 2021) despite their large population of speakers, have fewer resources than other languages of comparable speaker numbers. There is significantly less data available for AI systems to train on low-resource languages. To fill this gap, researchers have done work on such low-resource languages like Arabic (Mouheb et al., 2019), Indonesian (Ibrohim and Budi, 2018), Dravidian languages (Tula et al., 2022), Urdu and Roman-Urdu (Talpur et al., 2020), Turkish (Ccolltekin, 2020), and Vietnamese (Van Huynh et al., 2019), etc.

[29]https://en.wikipedia.org/wiki/Bengali_language

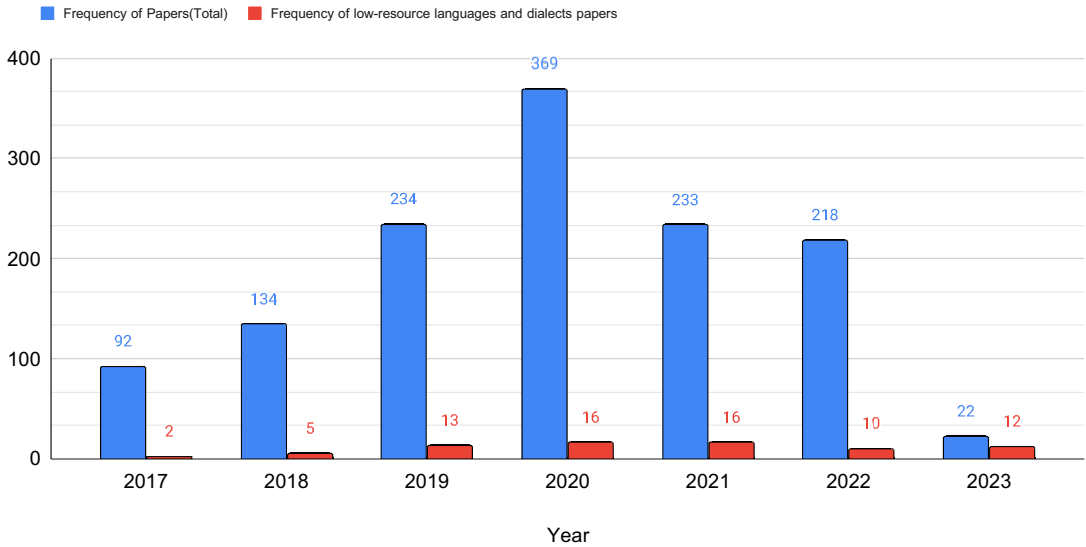

**Figure 2:** Numbers of relevant papers in years 2017-2023.

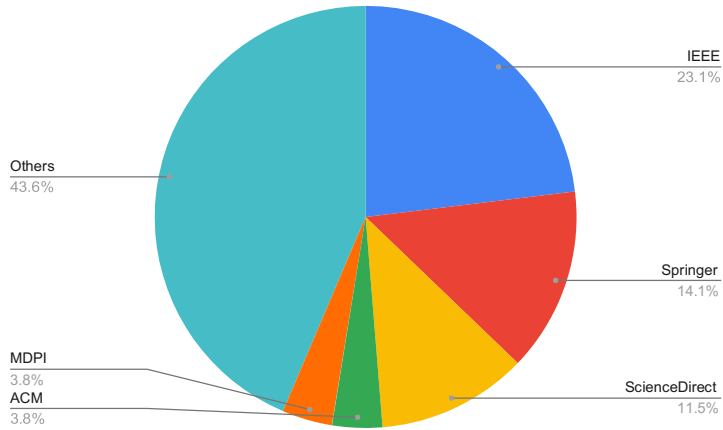

**Figure 3:** Statistics of papers gathered from various sources and databases.

Moreover, there is a collection of work that has been done on detecting cyberbullying and related content like abusive language (Mahmud et al., 2023), and hate speech (Djuric et al., 2015; MacAvaney et al., 2019; Romim et al., 2021). Researchers have been trying to solve this problem mainly in three ways. One is with traditional Machine Learning (ML) techniques (Mahmud et al., 2023) by applying classic Bag-of-Words language modeling and TF-IDF weighting schema for feature representation. The second is with transformer-based transfer learning models like BERT (Das et al., 2021), BERT-m (Das et al., 2021; Gaikwad et al., 2021), DistilmBERT, IndicBERT (Saha et al., 2021), and the third is with hybridization of various deep learning models like Simple RNN (Dewani et al., 2021), LSTM, BiLSTM (Badjatiya et al., 2017), GRU (Ishmam and Sharmin, 2019), CNN (Cao et al., 2020).

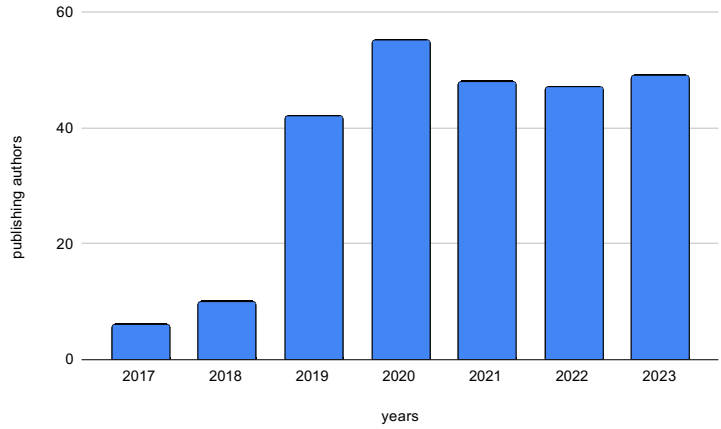

**Figure 4:** Increase in numbers of authors published in years 2017-2023.

**Table 2**
Numbers of papers authored (including co-authored) by the same authors, published in years 2017-2023.

| Numbers of authors | Number of papers per author |
|---|---|
| 199 | 1 |
| 22 | 2 |
| 1 | 3 |
| 2 | 4 |

**Table 3**
Languages covered in this survey with their respective language families.

| Studied Languages and Dialects | Language Family |
|---|---|
| Bangla,Hindi,Marathi,Nepali,Urdu,Sinhala | Indo-Aryan[19] |
| Tamil, Kannada, Malayalam | Dravidian[20] |
| Arabic,Tunisian Arabic,Egyptian Arabic,Algerian Arabic | Semitic[21] |
| Turkish | Oghuz[22] |
| Persian | Indo-Iranian[23] |
| Indonesian,Vietnamese | Austronesian[24] |
| Swahili,Zulu,Sesotho | Niger-Congo[25] |
| Afrikaans | Germanic[26] |
| Portuguese | Italic[27] |
| Slovenian | Balto-Slavic[28] |

The next sections summarize papers on automatic cyberbullying detection as well as papers closely related to this field such as abusive language detection and hate speech language detection, specifically done in low-resource languages(See Table 3).

### 3.2.1. Bangla Language

Cyberbullying detection is a classification problem, usually reduced to classifying an input into one of two classes ("cyberbullying", or "not cyberbullying"). Studies in this section focus mostly on such binary classification for Bangla language (Romim et al., 2021) in chronological order.

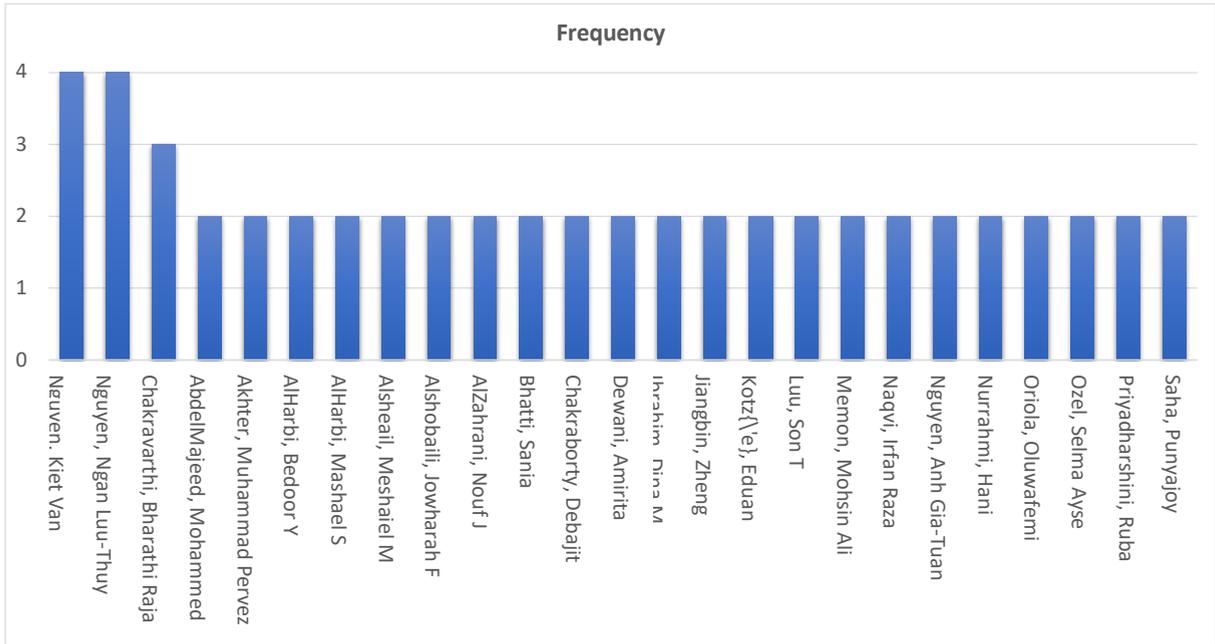

**Figure 5:** Names of papers authored (including co-authored) with more than one paper in years 2017-2023.

Tables 4 and 5 shows the recent studies on cyberbullying detection and related topics in Bangla. Eshan and Hasan (2017) tried Multinomial Nave Bayes, Random Forest, and SVM - a set of traditional machine learning classifiers - where they collected data by scraping the Facebook pages of popular celebrities. In this experiment, unigram, bigram, and trigram features were collected and TF-IDF vectorizers were used for weighting. They compared their methods on datasets of different sizes, namely, 500, 1,000, 1,500, 2,000, and 2,500 samples, respectively.

Hussain et al. (2018) proposed a root-level algorithm with unigram string features to detect abusive Bangla comments. They collected 300 comments from Facebook pages, Prothom Alo news and YouTube channels. They used 80% of their collected comments to train and 20% for testing. They divided their dataset into three sets of 100, 200, or 300 comments and tested their system with an average obtained result of 0.689.

Akhter et al. (2018) compared several machine learning-based classification models using a corpus of Bangla text that has been categorized as either expressing bullying or not across a variety of social media platforms. With a detection accuracy of 0.97 for Bangla text, a Support Vector Machine, or SVM-based method outperformed other models using a cross-validation experiment. The authors also suggest that their Bangla cyberbullying detection system's classification accuracy could be further enhanced by the impact of user-specific data, such as location, age, and gender.

Awal et al. (2018) used classic ML techniques, like Naive Bayes, to detect abusive comments. The authors collected 2,665 English comments from YouTube and then they translated them to Bengali in two ways. These were: i) Direct translation to Bangla. ii) Dictionary based translation to Bangla. Their proposed implementation reached 0.8057 accuracy.

Emon et al. (2019) applied different deep learning and machine learning-based algorithms with Count and TF-IDF vectorizers to detect abusive Bengali text. They collected 4,700 comments from SNS such as Facebook and Youtube as well as newspaper outlets like Prothom Alo online[30] and labeled these data in seven different classes i.e. slang, religious hatred, personal attack, politically violated, anti-feminism, positive, and neutral. They got the highest score of 0.8220 accuracy with the Recurrent Neural Network (RNN) algorithm.

Jahan et al. (2019) created a dataset by collecting comments from public Facebook sites such as news and celebrity pages using online comment scraping tools. Three machine learning methods were used: SVM, Random Forest, and Adaboost. Their method, the Random Forest classifier performed best in terms of accuracy and precision, with scores of 0.7214 and 0.8007, respectively, while Adaboost performed best in terms of recall, with a score of 0.8131.

---

[30]Prothom Alo is a daily newspaper published in Bangladesh in the Bengali language. https://www.prothomalo.com/

Ishmam and Sharmin (2019) annotated a dataset collected from Facebook into six classes. They included linguistic and quantitative features, as well as various text preprocessing techniques such as removing punctuation, bad characters, hashtags, URLs, mentions, tokenization, stemming, and so on. They applied neural networks like GRU with a few other ML classifiers and classified the data based on historical, religious, cultural, social, and political contexts.

Karim et al. (2021) used various machine learning classifiers (e.g., Logistic regression , SVM etc.) and deep NNs (i.e., CNN, Bi-LSTM etc.) for the detection of hate speech in the Bengali language. Using both ML and DNN they categorized their dataset from Facebook, YouTube comments, and newspapers into political, religious, personal, and geopolitical hate speech and got the F1 score of 0.78(political), 0.91(personal), 0.89(geopolitical) and 0.84(religious) for hate speech detection in Bengali.

Sazzed (2021) created a 3,000 transliterated Bengali comment corpus, with 1,500 abusive remarks and 1,500 non-abusive ones. They used supervised machine learning such as Logistic regression (LR), support vector machine (SVM), random forest (RF), and deep learning-based bidirectional long short-term memory (BiLSTM) for baseline assessments. The support vector machine (SVM) classifier performed the best at detecting abusive content.

Romim et al. (2021) applied neural networks like LSTM and BiLSTM in classifying hate speech in the Bengali language using word embeddings pre-trained with FastText, Word2Vec, and Glove. They introduced the largest to-date Bengali dataset and compared various combinations of deep learning models and word embeddings. In this study ,all of the deep learning models performed well in the experiment SVM had the greatest accuracy of 0.875.

To detect abusive comments, (Islam et al., 2021) employed a large dataset obtained from Facebook and YouTube and computing the best possible output using multinomial Nave Bayes (MNB), multilayer perceptron (MLP), support vector machine (SVM), decision tree, random forest, stochastic gradient descent (SGD), ridge, perceptron, and k-nearest neighbors (k-NN). The dataset was processed with a Bengali stemmer after randomly under sampling the dominant class. As a result, when applied to the entire dataset, SVM achieved the greatest accuracy of 0.88.

To classify the abusive comments on Facebook, Aurpa et al. (2022) applied transformer-based deep NN models like BERT (Devlin et al., 2019) and ELECTRA (Clark et al., 2020) and used 44,001 Facebook comments as a dataset. As for the outcome, their models reached 0.85 and 0.8492 test accuracy for classifier models of BERT and ELECTRA, respectively.

### 3.2.2. Hindi Language

Table 6 shows the recent studies on cyberbullying detection and related topics in Hindi.

Tarwani et al. (2019) studied the identification of cyberbullying in Hinglish, a language widely used in India. The authors developed the Hinglish Cyberbullying Comments labeled dataset, which includes comments from social media sites like Instagram and YouTube. To automatically identify instances of cyberbullying, They created eight distinct machine learning sentiment classification models. These models are assessed using performance metrics such as accuracy, precision, recall, and f1 score. The top eight baseline classifiers are finally integrated to create a hybrid model and performs better, with accuracy and F1-scores of 0.8026 and 0.8296, respectively.

Pawar and Raje (2019) developed a Multilingual Cyberbullying Detection System in Hindi and Marathi. The authors developed a prototype that worked with the datasets in these two languages. They conducted experiments to detect cyberbullying in these two languages using this prototype and experiments on a variety of datasets and showed that both languages have an accuracy of up to 0.97 and an F1-score of up to 0.96.

Sreelakshmi et al. (2020) studied hate speech detection in Hindi-English mix-coded language on social media sites. In this regard, they collected a total number of 10,000 data samples and used three ML methods, namely, SVM-linear, SVM-RBF and Random Forest. They used three experiments for these ML methods using fastText and the highest accuracy of 0.8581 was achieved by SVM-RBF.

Maity and Saha (2021) created a benchmark corpus to identify Hindi-English code-mixed language cyberbullying targeted against women and girls. The BERT, CNN, GRU, and capsule networks served as the foundation for the model they built. Different traditional machine learning models (SVM, LR, NB, and RF) were assessed as baselines alongside deep neural network models (CNN, LSTM). Their model (BERT+CNN+GRU+Capsule) surpassed the baselines with overall accuracy, precision, recall, and F1-measure values of 0.7928, 0.7867, 0.8199, and 0.8030, respectively.

Mehendale et al. (2022) used NLP and ML to develop a model that could identify disparaging or derogatory terms in both English and Hinglish (a code-mixed version of English with Hindi vocabulary).They used SVM,RF,KNN and LR to produce better outcomes. Finally they achieved 0.95 accuracy.

Sharma et al. (2022) introduced MoH or (Map Only Hindi),which means Love in Hindi. The MoH refers to a word-level transliteration pipeline that recognizes language tags and performs partial transliteration of code-switched text while maintaining sentence meaning. After that, the researchers tweaked Multilingual Bidirectional Encoder Representations from Transformers (M-Bert) and Multilingual Representations for Indian Languages (MuRIL) to investigate if Devanagari Hindi-English code-switching is adequate for Bert-based models. At first, researchers examined the performance of 'MoH' mapped text using standard machine learning models and reported an average 0.13 boost in F1 scores. Furthermore, when compared to baseline models, their proposed method outperformed them by 0.06.They also created a 'MoH' technique with multiple data simulations using the present translation library. In this case, 'MoH' outperformed the others by 0.15.

**Table 4**
The recent studies on low-resource Bangla cyberbullying detection or related topics[**Accuracy(A),Precision(P),F1 Score(F1)**]

| Reference | Classifier Model | Highest Score | Language for Model | Dataset Size | Different Label | Origin Sources |
|-----------|------------------|---------------|--------------------|--------------|-----------------|----------------|
| Eshan and Hasan (2017) | MNB RF SVM | 0.8(A) | Bangla | 2,500 | - | Facebook, |
| Hussain et al. (2018) | Root-Level Algorithm | 0.689(A) | Bangla | 300 | Bully, Not Bully | Facebook YouTube News portal |
| Akhter et al. (2018) | SVM,NB DT,KNN | 0.97(A) | Bangla | 2,400 | Bullied(10%) Not bullied | Facebook Twitter |
| Awal et al. (2018) | NB | 0.8057(A) | Bangla | 2,665 | Abusive(45.55%), Non abusive | YouTube |
| Emon et al. (2019) | LinearSVC, LR (Logit), MNB, RF,ANN, RNN+LSTM | 0.822(A) | Bangla | 4,700 | Slang(19.57%), Religious Hatred(13.15%), Personal attack(12.36%), Politically violated(13.28%), Anti-feminism(0.87%), Positive, Neutral. | Facebook, YouTube, News portal |
| Jahan et al. (2019) | SVM, Random Forest, Adaboost | 0.7214(A) 0.8(P) | Bangla | 2,000 | Abusive(78.41%), Non abusive | Facebook |

Anand et al. (2023) performed multilingual offensive language detection operations using deep learning to categorize the offensive language. Features for segmented data were selected using a fuzzy convolutional neural network (FCNN) and feature extraction and classification were carried out using the ensemble architecture consisting of a BiLSTM model with a hybrid Naive Bayes and support Vector Machines (SVM).

Madhu et al. (2023) created a dataset to classify hate speech in code-mixed Hindi-English conversations. Their applied classification pipeline consisted of an improved SentBERT and an LSTM acting as the classifier. On the provided dataset, this pipeline achieved a macro-average F1 score of 0.892. The models with the highest performance contained KNN, SentBERT, and ABC weighting-based separate pipeline, with a slightly lower performance of 0.807.

### 3.2.3. Dravidian Languages

Table 7 shows the recent studies on cyberbullying detection and related topics in Dravidian languages.

(Sharif et al., 2021) combined word embedding and tf-idf features along with models from SVM, LR, LSTM, and LSTM+Attention. The results demonstrated that the ML ensemble outperformed DL approaches in terms of accuracy. The weighted f1 scores of Tamil, Malayalam, and Kannada all increased from 0.73 to 0.76, 0.88 to 0.93, and 0.48 to 0.71, respectively and finally evaluated the Transformer-based models overall performance using the ensemble method.

**Table 5**
The recent studies on low-resource Bangla cyberbullying detection or related topics[**Accuracy(A),Precision(P),F1 Score(F1)**]

| Reference | Classifier Model | Highest Score | Language for Model | Dataset Size | Different Label | Origin Sources |
|---|---|---|---|---|---|---|
| Ishmam and Sharmin (2019) | GRU SVC,LSVC RF,NB | 0.7010(A) | Bangla | 5,126 | Hate speech (19.2%), Inciteful (10.77%), Religious comment (14.9%), Religious hatred (15.68%), Political comment (23.43%), Communal hatred (15.67%) | Facebook |
| Karim et al. (2021) | LR, SVM CNN, BI-LSTM Conv-LSTM BERT | 0.78(F1) 0.91(F1) 0.89(F1) 0.84(F1) | Bangla | 8,087 | Personal (43.44%), Political (12.35%) Religious (14.97%), Geopolitical (29.23%) | Facebook |
| Sazzed (2021) | SVM,LR RF,BiLASTM | 0.827(F1) | Bangla | 3,000 | Abusive(10%), Non abusive | YouTube |
| Romim et al. (2021) | LSTM,BiLSTM | 0.8750(A) | Bangla | 30,000 | Hate speech (33.33%), Not hate speech | Facebook, YouTube, |
| Islam et al. (2021) | MNB, MLP, SVM, DT,RF, SGD,k-NN | 0.88(A) | Bangla | 9,760 | Abusive(50%), Non abusive | Facebook YouTube |
| Aurpa et al. (2022) | transformer-based DNN, BERT, ELECTRA | 0.85(A) (BERT), 0.8492(A) (ELECTRA) | Bangla | 44,001 | Sexual (20.29%), Not Bully (34.86%) , Troll(23.78%), Religious (17.22%), Threat(3.85%) | Facebook |

(Saha et al., 2021) worked on low-resourced Dravidian languages Tamil, Kannada and Malayalam and collected a dataset for these languages from YouTube comments and posts. They used classification models as well as transformer-based models to detect the best performance in finding hate speech in social media. The overall test performance measurements show that XML-RoBERTa (XMLR-large) in Tamil gives the best F1 score of 0.78, Custom XLM-RoBERTa-base (XLMR-C) in Kannada gave the best F1 score of 0.74 and for Malayalam both the models had the best accuracy of overall F1 score of 0.97.

**Table 6**
The recent studies on cyberbullying detection and related topics in Hindi [**Accuracy(A),Precision(P),F1 Score(F1)**]

| Reference | Classifier Model | Highest Score | Language for Model | Dataset Size | Different Label | Origin Sources |
|---|---|---|---|---|---|---|
| Tarwani et al. (2019) | machine learning models | 0.8026(A) 0.8296(F1) | Hinglish | - | | Instagram, YouTube, |
| Pawar and Raje (2019) | SGD,MNB,LR | 0.97(A) 0.96(F1) | English, Hindi, Marathi | 1,431 | Bullying(9%) Non Bullying | formspring .me, movie reviews, tour reviews, newspaper reviews, |
| Sreelakshmi et al. (2020) | SVM-linear, SVM-RBF Random Forest. | 0.8581(A) | Hindi-English | 10,000 | Hate(50%), Non Hate | Twitter, |
| Maity and Saha (2021) | SVM, LR, NB, RF, CNN, LSTM, BERT+CNN +GRU+Capsule | 0.7928(A), 0.7867(P), 0.8199(R) 0.8030(F1) | Hindi-English | 5,062 | Bully(51.48%) Non Bully | Facebook |
| Mehendale et al. (2022) | SVM,RF, KNN,LR | 0.95(A) | English-Hinglish, | 18,000 | Bullying(64%) Non bulying | Twitter, Whatsapp, YouTube, |
| Sharma et al. (2022) | NB,LR,SVM, RF,XGBoost GradientBoosting, AdaBoost, MoH + M-Bert, MoH + MuRIL | MoH outperforms the rest by 0.15(F1) | Hindi-English | 12,000, 4,575, 3,679 | Non-Offensive Abusive(52.78%) Hate Inducing(8.78%) Hate Speech (36.31%) Normal Speech, Overtly Aggressive(39.25%), Covertly Aggressive(39.26%), Non-Aggressive | Facebook, Twitter |
| Anand et al. (2023) | Ensemble of BiLSTM SVM+Naïve Bayes | 0.98(A) | Hindi | 2,963 | Offensive(28.59%) NOT Hate(7.9%) Profane(4.99%) None | YouTube Twitter Facebook |
| Madhu et al. (2023) | LR,NB KNN,SVM CNN,LSTM | 0.892(F1) | Code-mixed Hindi | 7,088 | Hate offensive (49.88%) NOT hate offensive | Twitter |

(Shibly et al., 2022) developed a system to detect hate speech in Tamil-English code-mixed language and collected a dataset of 35,442 tweets and divided them into offensive and not offensive groups. The ML algorithms they used in this research were SVM, NB, Logistic Regression (LR), Decision Tree (DT), Random Forest (RF), Gradient Boost (GB) and K Nearest Neighbor (KNN). NB obtained the highest F1 score, but SVM has obtained the highest accuracy in hate speech detection using 10-fold Cross Validation. The average accuracy of the ensemble classification models obtained was 0.9067.

**Table 7**
Recent studies on low-resource Dravidian languages cyberbullying detection or related topics[**Accuracy(A),Precision(P),F1 Score(F1)**]

| Reference | Classifier Model | Highest Score | Language for Model | Dataset Size | Different Label | Origin Sources |
|---|---|---|---|---|---|---|
| Sharif et al. (2021) | SVM, LR, LSTM, LSTM+Attention. m-BERT, Indic-BERT, XLM-R | 0.76(F1), 0.93(F1), 0.71(F1) | Tamil, Kannada, Malayalam | 33,204 66,265 44,415 | Not offensive, Offensive-Targeted-Insult-Other, Insult-Individual, Insult-Group, Not- Tamil, Not-Malayalam, Not-Kannada, Offensive-Untargetede (4.2%) | Social media |
| Saha et al. (2021) | XLMR-base(A) XLMR-large XLMR-C (B) mBERT-base(C) IndicBERT MuRIL DistilBERT | 0.78(F1) 0.74(F1) 0.97(F1) | Tamil, Kannada, Malayalam | 43,919, 7,772, 20,010 | Not-offensive, Offensive-untargeted(8.27%), Offensive-targeted-Individual(6.75%), Group(7.12%), Other(1.34%), Not-in-indented-language | YouTube |
| Shibly et al. (2022) | SVM, NB, LR, DT,RF, GB,KNN | 0.9067(A) | Tamil-English, | 35,442 | Offensive, Not offensive | YouTube |
| Subramanian et al. (2022) | Naïve Bayes, SVM, LR K- NN mBERT, MuRIL XLM-RoBERTa | 0.885(A) | Tamil | 6,534 | Offensive(19.68%), Not offensive, Not Tamil | YouTube |
| Balouchzahi et al. (2022) | n-gram-MLP classifier 1D Conv-LSTM n-gram-MLP trained with n-grams | 0.56(F1) 0.43(F1) | Tamil Code mixed | 2,800, 7,279 | Misandry, Counter-speech, Xenophobia, Hope-Speech, Misogyny, Homophobia, Transphobic, Not-Tamil | online platforms |
| Chakravarthi et al. (2023) | SVM,NB,RF,DT LGBM,EWDT, EWODT MPNet,CNN,BERT | 0.85(F1) 0.98(F1) 0.76(F1) | Tamil Kannada Malayalam | 43,919 20,010 7,771 | Offensive(23.49% 3.5%,18.98%), Not offensive | YouTube |
| Rajalakshmi et al. (2023) | SVM,SGD,DT IndicBERT,MuRIL | 0.86(A) | Tamil | 6,541 | Offensive(19.7%), Not offensive | YouTube |

**Table 8**
Recent studies on low-resource Sinhala language cyberbullying detection or related topics [**Accuracy(A),Precision(P),F1 Score(F1)**]

| Reference | Classifier Model | Highest Score | Language for Model | Dataset Size | Different Label | Origin Sources |
|---|---|---|---|---|---|---|
| Sandaruwan et al. (2019) | MNB, SVM, Random Forest Decision Tree | 0.9233(A) | Sinhala | 3,000 | Hate(33.33%) Offensive(33.33%) Neutral(33.33%) | Facebook, YouTube |
| Amali and Jayalal (2020) | NB, KNN, SVM | 0.91(F1) | Sinhala | 625 | Cyberbullying(10%) Non-cyberbullying | Twitter |

(Vasantharajan and Thayasivam, 2022) initiated a system based on various techniques and neural network models to detect offensive language in social media in Tamil language mix-coded. They used mBERT, DistilBERT, XLM-RoBERTa, CNN-BiLSTM, ULMFiT models in their experiment. As a result, the best accuracy rate was found in the ULMFiT model that scored 0.7287 precision with a 0.7401 F1 score.

(Subramanian et al., 2022) studied the detection of offensive language in a low-resourced Tamil language used in YouTube comments. They used adapter-based model to get better accuracy. In this experiment, XML-RoBERTa (large) achieved the best accuracy of 0.885.

(Balouchzahi et al., 2022) studied "Abusive Comment Detection in Tamil- ACL 2022" to categorize the various types of abusive language that can be found in texts written in both code-mixed and native Tamil script. The n-gram-MLP classifier outperformed the other two models1D Conv-LSTM and n-gram-MLP trained with n-grams on texts written in both code-mixed and native Tamil scripts, with average weighted F1-scores of 0.560 and 0.430, respectively.

Chakravarthi et al. (2023) proposed a multilingual MPNet and CNN fusion model for low-resource Dravidian languages (Tamil, Malayalam, and Kannada) to categorize code-mixed social media comments and posts as either offensive or non-offensive. The model beat the baseline models in recognizing offensive language, attaining weighted average F1-scores of 0.85, 0.98, and 0.76, for the respective languages, and it performed 0.02, 0.02, and 0.04 better than the baseline models ensembles with Decision Tree (EWDT) and ensemble without Decision Tree (EWODT) for Tamil, Malayalam, and Kannada, respectively.

Rajalakshmi et al. (2023) proposed TF-IDF and the pre-trained transformer models BERT, XLM-RoBERTa, IndicBERT, mBERT, TaMillion, and MuRIL as suitable embedding techniques for Tamil text representation. They used a variety of classifiers like ensemble learning models, Decision Trees, SVM with stochastic gradient descent, and logistic regression. The methods were applied in a multilingual word embedding model MuRIL and the downstream classifier applied as a majority vote ensemble achieved the highest performance on stemmed data. The final best proposed model recognized offensive content with an accuracy of 0.86 of the Tamil YouTube comments and an F1 score of 0.84.

### 3.2.4. Sinhala Language

Table 8 shows the recent studies on cyberbullying detection and related topics in the Sinhala language.

Sandaruwan et al. (2019) studied the detection of hate speech in social media using ML models in the low-resourced language of Sinhala. The corpus of this language contained 3,000 comments from Facebook and YouTube in the Sinhala language. The models they used in this experiment were Multinomial Naïve Bayes (MNB), SVM, and Random Forest Decision Tree (RFDT). Among these classifier models, MNB obtained the highest accuracy of 0.9233 and 0.84 of Recall value.

Amali and Jayalal (2020) initiated a hybrid technique, which combined a rule-based strategy and machine learning, to recognize cyberbullying remarks on social networking sites in the Sinhala language. A dataset of 625 tweets that had been manually labeled as offensive or not offensive was generated by the algorithm. After that, they used the Naive Bayes, KNN, and Support Vector Machine algorithms. The highest F1-score of 0.91 was obtained by SVM with an RBF kernel.

### 3.2.5. Marathi Language

Table 9 shows the recent studies on cyberbullying detection and related topics in Marathi language.

**Table 9**
The recent studies on low-resource Marathi and Napali languages cyberbullying detection or related topics[Accuracy(A),Precision(P),F1   Score(F1)]

| Reference | Classifier Model | Highest Score | Language for Model | Dataset Size | Different Label | Origin Sources |
|---|---|---|---|---|---|---|
| Gaikwad et al. (2021) | XLM-R Zero- and Few-shot learning, | 0.8461(F1) | Marathi | 2,499 | Offensive(35.05%) Non-offensive | Twitter |
| Glazkova et al. (2021) | XLM-RoBERTa LaBSE | 0.8808(F1) | Marathi | 2,499 | Hate, Offensive, Profane (35.7%) | Twitter |
| Niraula et al. (2021) | SVM,RF, LR M-BERT | 0.87, 0.71, 0.45, 0.01 (F1) | Nepali text Devanagari | 7,248 | Sexist(1.18%), Racist(3.6%), Other -Offensive(34.2%), Non-offensive(61%) | Twitter, YouTube, Facebook, Blogs, News Portals |

Gaikwad et al. (2021) created the first dataset for the low-resource Marathi language (MOLD). They experimented with the cross-lingual transformer architecture to project prediction in Marathi from English and two Indo-Aryan languages, Bangla and Hindi. They applied both monolingual models as well as multilingual models in cross-lingual transfer learning, especially zero- and few-shot learning, which boosted the performance.

(Glazkova et al., 2021) employed two tasks that required the binary and fine-grained categorization of English tweets that contain hate, offensive, and profane content, in addition to a task involving the identification of problematic content in Marathi. To discriminate between comments that are hate, profane and offensive, they employ a one-vs-rest technique based on Twitter-RoBERTa. Their models came in second and third with F1-scores of 0.6577 for English subtask B and 0.8199 for English subtask A, respectively. Based on the Language-Agnostic BERT Sentence Embedding, they offer a remedy for the Marathi problems .Their developed model finished Marathi subtask A's second result with an F1 of 0.8808.

### 3.2.6. Nepali Language

Table 9 represents the recent studies on cyberbullying detection and related topics in Nepali language.

(Niraula et al., 2021) highlighted the problems with how material from Nepali social media is analyzed and defines abusive language in Nepali, a language with few resources. They gathered a variety of social media postings in Nepali text Devanagari and manually annotated 7,248 of them to create a labeled data collection. In addition, they used a variety of traditional machine learning and modern deep learning models and features for the detection of offensive language and describe the first baseline approaches for identifying profanity in Nepali.

### 3.2.7. Urdu Language

Table 10 represents the recent studies on cyberbullying detection and related topics in Urdu language.

Akhter et al. (2020) studied automatic offensive language detection in Roman Urdu and Urdu, where the dataset was based on YouTube comments. A total of seventeen models were used in this experiment, namely, NB, BayesNet, k-NN, Hoeffding Tree, J48, REPTree, SVM-Linear, SVM-Radial, SVM-Sigmoid, SVM-Polynomial, RF, RT, LR, LogitBoost, SimpleLogistic, OneR, and JRip. Among all of these models, LogitBoost had the best accuracy of 0.992 for the Roman Urdu language. On the other hand, SimpleLogistic had the best accuracy of 0.958 score for the Urdu language.

To detect the cyberbullying Roman Urdu behavior, Talpur et al. (2020) created a model based on a lexicon with features derived from Twitter and compared it with supervised machine learning model. According to an evaluation model, the developed model produced results calculated as the area under the receiver operating characteristics curve (AUC) of 0.986, with an F1-measure of 0.984.These findings imply that the proposed lexicon-based method produced sufficiently usable results.

**Table 10**
The recent studies on low-resource Urdu language cyberbullying detection or related topics [**Accuracy(A), Precision(P), F1 Score(F1)**]

| Reference | Classifier Model | Highest Score | Language for Model | Dataset Size | Different Label | Origin Sources |
|---|---|---|---|---|---|---|
| Akhter et al. (2020) | NB, BayesNet, k-NN, Hoeffding Tree, J48, REPTree, SVM-Linear, SVM-Radial, SVM-Sigmoid, SVM-Polynomial, RF, RT, LR, LogitBoost, SimpleLogistic, OneR,JRip. | 0.992(A), 0.958(A) | Roman Urdu, Urdu | 10,000, 2,000 | Offensive( 50%), Non offensive ( 50%) | YouTube |
| Rizwan et al. (2020) | BERT+CNNgram | 0.82(A) | Roman Urdu | 10,000 | Abusive /Offensive(23.99%) Sexism(8.38%) Religious Hate(7.81%) Profane(0.64%) Normal(53.43%) | Twitter |
| Das et al. (2021) | XGboost, LGBM, m-BERT | 0.88(F1), 0.54(F1) | Urdu | 3,500, 9,950 | Abusive(50%) Non-Abusive Threatening(17.99%) Non-Threatening | Twitter |
| Dewani et al. (2021) | RNN-LSTM, RNN-BiLSTM CNN | 0.8550(A) | Roman Urdu | - | Bullying No Bullying | Twitter |
| Akhter et al. (2022) | NB, SVM IBK, LR JRip CNN, LSTM, BLSTM, CLSTM | 0.914(A), 0.962(A) | Roman Urdu, Urdu | 10,000, 2,171 | Abusive(51.08%,50%) Non Abusive | YouTube |
| Dewani et al. (2023) | SVM, MNB LR,DT Ada boost XGBoost, Bagging classifier | 0.83(A) SVM | Roman Urdu, | - | Cyberbullying Non cyberbuulying | Twitter |
| Saeed et al. (2023) | SVM,LR, CNN,LSTM BERT | 0.86,0.8 0.81,0.72 (Macro-F1) | Urdu | 24,170 | Offensive(64.14%) Non Offensive | Twitter |
| Akram et al. (2023) | NB,SVM LR,RF CNN,LSTM BILSTM,BERT | 0.834(F1) | Urdu | 21,759 | Hateful(39.76%), Neutral | Twitter |

Rizwan et al. (2020) created a dataset in Roman Urdu (RU) that was tagged with five fine-grained descriptors to detect the hate speech in social networking sites posts and constructed an annotated dataset named RUHSOLD that contained 10,012 RU tweets and included both fine-grained and coarse-grained classes of hateful speech and abusive language. They also offered a vocabulary of RU hateful terms. They looked through and gathered 50,000 new tweets using this revised terminology. Around 10,000 tweets were randomly selected for annotations from this updated Twitter repository. In terms of how well the several iterations of the proposed model performed, the BERT+CNNgram had the highest F1-score (0.75) and Accuracy (0.82).

Das et al. (2021) worked on the Urdu language. They divided the task into two different sub-tasks: one was focused on abusive language and the other was on threatening language detection. They used some machine learning and a few transformer-based pre-trained models. One of the m-BERT-based models, referred to as the "dehatebert-mono-arabic", exploited the similarities between the Urdu language and the Arabic language.

Dewani et al. (2021) studied cyberbullying detection of social media posts in Roman Urdu, which is a highly under-resourced language. They used RNN-LSTM, RNN-BiLSTM and CNN models for the experiment using Roman Urdu tweets as a dataset. Results showed that RNN-LSTM and RNN-BiLSTM had the best accuracy of 0.855 and 0.85 respectively. Also, the F1 scores were 0.7 and 0.67 respectively over the aggression class.

Similarly, Akhter et al. (2022) performed abusive language detection from social media comments in Roman Urdu and Urdu languages. In this regard, they have collected 10,000 Roman Urdu texts from GitHub and 2,000 Urdu comments from Urdu Offensive Dataset (UOD). They used 5 ML models (e.g., NB, SVM, IBK, Logistic Regression, and JRip) and four DL models (e.g., CNN, LSTM, BLSTM, and CLSTM) to classify abusive comments in the provided dataset. As a result, they achieved 0.962 accuracy on Urdu and 0.914 accuracy on Roman Urdu.

Dewani et al. (2023) suggested using ensemble approaches based on voting, and used machine learning algorithms for recognizing cyberbullying in Roman Urdu micro-texts. The BOW model with TFIDF weighting, word N-grams, combinations of N-grams, and statistical features were all extracted using GridSearchCV and cross-validation techniques and the identification method was created to reduce user-provided text input while yet being flexible and, in particular, taking into account users' varied vocabulary on social media.

Saeed et al. (2023) focused on identifying offensive speech, hate speech, and various levels of hate speech in Urdu based on religion, racism, and national origin. They categorized different types of hate speech into levels based on the intensity of the expressed hatred.

Akram et al. (2023) created a corpus (ISE-Hate) for identifying hate content in Urdu and examined eight supervised machine learning and deep learning methods for automated hateful content identification. Finally, the method based on Bidirectional Encoder Representation with Transformers (BERT) achieved the highest score for detecting hateful content in Urdu tweets.

Arshad et al. (2023) created an Urdu hate lexicon and annotated 7,800 Urdu tweets. They proposed transfer learning with multilingual BERT embeddings in FastText Urdu word embeddings (RoBERTa). Five deep learning architectures had their performance evaluated in contrast to more well-known machine learning models such as decision trees, support vector machines, naive bayes, logistic regression, and ensemble models. They showed that RoBERTa performed better in multi-class classification issue than baseline deep learning and machine learning models, reaching a macro F1-score of 0.82.

### 3.2.8. Arabic Language and Dialects

Table 11 shows the recent studies on cyberbullying detection and related topics in Arabic language.

In order to identify Arabic cyberbullying, Haidar et al. (2018) proposed a solution that made use of deep learning techniques and made an Arabic dataset for Health Services from Tweets. A Feed Forward Neural Network (FFNN) was created specifically for the recognition of cyberbullying in Arabic. To get the highest accuracy, the FFNN underwent several iterations of training with various parameter adjustments. It obtained the best accuracy of 0.9352, Test Accuracy of 0.9333, and Network Validation Accuracy of 0.9456.

The first attempts for the Arabic language, Mouheb et al. (2019) used machine learning to automatically detect bullying keywords using the Naive Bayes classifier. They determined the likelihood that a comment will be interpreted as bullying. The precision of the Nave Bayes Classification method was 0.959.

AlHarbi et al. (2019) established a cyberbullying lexicon before obtaining data for research using the PMI, Chi-square, and Entropy methodologies using Twitter API, Microsoft-Flow, and YouTube comments. They recommended using sentiment analysis and lexicon approaches to automatically detect cyberbullying. The results show that the PMI approach outperformed the Chi-square and Entropy strategies in detecting cyberbullying.

AlHarbi et al. (2020) used machine learning algorithms for detecting cyberbullying in English and sentiment analysis in Arabic on social media in which they analyzed the offensive tweets collected from social media sites. By comparing the accuracy of the classifications, they came to the conclusion that Ridge Regression (RR) and Logistic Regression (LR) had the highest accuracy, where RR had 0.99 accuracy in sentiment analysis of Arabic text.

**Table 11**
The recent studies on low-resource Arabic language cyberbullying detection or related topics[**Accuracy(A),Precision(P),F1 Score(F1)**]

| Reference | Classifier Model | Highest Score | Language for Model | Dataset Size | Different Label | Origin Sources |
|---|---|---|---|---|---|---|
| Haidar et al. (2018) | FFNN | 0.9352(A) | Arabic | 39,803 | Bullying (7.57%) Non Bullying | Twitter |
| Mouheb et al. (2019) | NB | 0.959(P) | Arabic | 25,000 | Bullying Non Bullying | YouTube Twitter. |
| AlHarbi et al. (2019) | PMI, Chi-square, Entropy | 0.81(F1) | Arabic | 100,327 | Bullying Non Bullying | Twitter YouTube |
| AlHarbi et al. (2020) | RR LR | 0.99(A) | Arabic | - | Bullying Non Bullying | Social Media |
| Alsafari et al. (2020) | CNN, LSTM, GRU, SVM, NB,LR | 0.8516(F1) | Arabic | 5,340 | Hateful/ Offensive (35%) Clean(65%) | Twitter |
| Faris et al. (2020) | CNN, LSTM | 66.56(A) | Arabic | 3,696 | Positive Negative (22.8%) | Twitter |
| Nayel (2020) | LinearSVM MLP | 0.9017(F1) | Arabic | 8,000 | Offensive (19.86%) Not Offensive | Twitter |
| Alshalan and Al-Khalifa (2020) | CNN, GRU CNN+GRU, BERT. | 0.79(F1) | Arabic | 8,964 | Hate (28.32%) Non Hate | Twitter |
| Alduailaj and Belghith (2023) | SVM NB | 0.95742(A) SVM 0.70942(A) NB | Arabic | 30,000 | Bullying Non bullying | Twitter |
| Elzayady et al. (2023) | AraBERT | 0.823 Macro-F1 | Arabic | 6,839 | Hate(5.12%) Not hate | Twitter |
| Fkih et al. (2023) | LR,DT SVM,KNN RF,NB,NN | 0.90 Macro-F1 | Arabic | 6,964 | OFF(19%) NOT OFF HS(5%) NOT HS | Twitter |

Alsafari et al. (2020) have established a reliable method of detecting hateful and abusive speech in Arabic in addition to amassing a sizable corpus with a total of 800,000 entries. They extracted various n-gram features from the dataset, including unigrams, word-ngrams, char-ngrams, word/char-ngrams, and used word embeddings weighted with TF-IDF to assess their degree of importance throughout the corpus. They also directed pattern tests and applied a few deep learning approaches like CNN, LSTM, GRU, SVM, NB, and LR along with a variety of feature selection techniques. With an F1-macro of 0.8516, Precision of 0.8351, and Recall of 0.8819 in the SVM model, the word and char- ngrams conjunctions achieved the greatest accuracy.

For the automatic identification of cyberbullying, Faris et al. (2020) proposed a deep learning-based approach. Particularly, they focused on abuse spread on Twitter in the Arabic-speaking region. There were 3,696 tweets that were gathered in total. The word embedding characteristics were the input for the detection technique. The Word2Vec and AraVec techniques were used as word embeddings. When comparing Word2Vec and AraVec, it was shown that AraVec achieved better results across the board. Consequently, the Accuracy rate (0.66564), Recall (0.79768), Precision (0.68965), and F1 measure (0.71688) were obtained.

Nayel (2020) proposed a system designed to automatically detect the offensive language in Arabic tweets that was proposed for "SemEval-2020 Task12". They used SGD to create a linear classifier that relies on the TF-IDF vector space model, transforms the Twitter post into a vector, and then uses the vector space to perform linear classification. Their model produced an F1-score on the training dataset and testing dataset of 0.8420 and 0.8182, respectively. The system that performed the highest and the system in last place on the test set, respectively, recorded F1 scores of 0.9017 and 0.4451.

Alshalan and Al-Khalifa (2020) tried use ML models and deep neural network models to detect hate speech in the Saudi Twitter-sphere. They used a dataset of 9,316 annotated tweets and evaluated them with four ML models, namely, CNN, gated recurrent units (GRU), CNN+GRU, and BERT. Among these models, the CNN had the best accuracy with 0.79 of F1 score.

Alduailaj and Belghith (2023) proposed a model to identify cyberbullying in the Arabic language using machine learning. In the research, they used datasets gathered from YouTube and Twitter to train and evaluate the Support Vector Machine (SVM) classification algorithm, which was used to identify cyberbullying. Finally, they made use of Farasa[31] python library to further enhance cyberbullying detection.

Elzayady et al. (2023) created the first automatic system that can recognize hate speech in Arabic based on personality traits using a unique two-staged method for finding hate speech in Arabic social media. The first stage was based on searching for the presence of elements of hate speech, while the second stage consisted of ML-based hate speech detection.

Fkih et al. (2023) created a technique for finding offensive Arabic language content on social media platforms. They report that it was especially difficult to set up linguistic and semantic rules for slang modeling in Arabic. To overcome these issues, they provided a practical solution based on statistics and an AI-based technique. From the applied classifiers, Random Forest (RF) surpassed other well-known machine learning techniques in an experimental comparison with an accuracy of 90%.

Table 12 shows the recent studies on cyberbullying detection and related topics in Tunisian, Egyptian, and Algerian Arabic dialects.

Haddad et al. (2019) proposed T-HSAB, the first publicly available Tunisian dataset for hate and abusive speech, as a reference point for the automatic detection of offensive Tunisian content online. To ensure accurate dataset annotation, They thoroughly examined the data collection methods and the methods to develop the annotation standards. An extensive investigation of the annotations using the metrics of Cohen's Kappa and Krippendorff's alpha for annotation agreement ensured the consistency of the annotations.

To detect cyberbullying on social media, Farid and El-Tazi (2020) proposed an efficient strategy based on Arabic sentiment analysis. Arabic datasets for sentiment analysis are scarce because there has not been much research done on the subject compared to English. This is because Arabic is a distinctive and challenging language. In this study, a collection of Egyptian Arabic dialects, Modern Arabic, and emojis were compiled for the most accurate detection of cyberbullying. When analyzing positive hashtags, the suggested algorithm performed better than when analyzing negative ones, maintaining an accuracy score of at least 0.73 for these bullying-related hashtags.

Boucherit and Abainia (2022) focused on the Algerian dialect of Arabic to uncover offensive Facebook remarks. Then they simulated different dialects and languages like Berber, French, English, the roman and Arabic scripts (i.e. Arabizi). They created a new corpus with more than 8.7k texts that have been manually categorized as normal, abusive, and offensive due to the dearth of works on the same language. Several tests used the contemporary text classification classifiers BiLSTM, CNN, FastText, SVM, and NB. The outcomes revealed good results.

### 3.2.9. Turkish Language

Table 13 shows the recent studies on cyberbullying detection and related topics in Turkish language.

Ozel et al. (2017) studied the detection of cyberbullying in Turkish social media messages and created a dataset from Turkish-language Instagram and Twitter messages and used machine learning techniques such as Support Vector Machines (SVM), Decision Trees, Multinomial Naive Bayes, and k-Nearest Neighbors (kNN) classifiers. To improve classifier accuracy, they also employed information gain and chi-square feature selection methods. Cyberbullying detection improved when both words and emoticons in text messages were considered as features. In terms of classification accuracy and running time, the Multinomial Naive Bayes classifier was the most successful. When feature selection was used on the dataset, classification accuracy improved to 0.84.

---

[31]https://farasa-api.qcri.org/lemmatization

**Table 12**
The recent studies on low-resource Egyptian , Algerian and Tunisian Arabic Dialect cyberbullying detection or related topics[**Accuracy(A),Precision(P),F1 Score(F1)**]

| Reference | Classifier Model | Highest Score | Language for Model | Dataset Size | Different Label | Origin Sources |
|---|---|---|---|---|---|---|
| Haddad et al. (2019) | NB,SVM | 0.929(A) 0.777(A) | Tunisian Arabic | 6,075 | Normal, Abusive (18.66%), Hate (17.85%) | Social media |
| Farid and El-Tazi (2020) | SA,SATH, SATHE | 0.73(A) | Egyptian Arabic | - | Bully Not Bully | Twitter |
| Boucherit and Abainia (2022) | BiLSTM, CNN,FastText SVM,MNB GNB | 0.716,0.752(A) | Algerian Arabic | 8.7K | Normal, Abusive (13%), Offensive (31.46%) | Facebook |

**Table 13**
The recent studies on low-resource Turkish language cyberbullying detection or related topics[**Accuracy(A),Precision(P),F1 Score(F1)**]

| Reference | Classifier Model | Highest Score | Language for Model | Dataset Size | Different Label | Origin Sources |
|---|---|---|---|---|---|---|
| Ozel et al. (2017) | SVM,DT NB,KNN | 0.84(A) | Turkish | 900 | Bully(50%) Not Bully | Instagram Twitter |
| Bozyigit et al. (2019) | YSA2 | 0.91(F1) | Turkish | 3,000 | Positive Negative(50%) | Twitter |
| Ccolltekin (2020) | SVM | 0.773(F1) | Turkish | 36,232 | Offensive(19.4%) Non-offensive Targeted(20.3%) Not Targeted | Twitter |
| Beyhan et al. (2022) | BERTurk | 0.77(A) 0.71(A) | Turkish | 1,206, 1,278 | Not Hate Speech,Insult Exclusion Whishing Harm Threatening Harm (49.04%) | Twitter |
| Coban et al. (2023) | LR, MNB, SVM CNN, RNN, BERT | 0.928 Macro F1 | Turkish | 919,708 | Cyberbullying (11.06%) Non cyberbullying | Facebook |

Bozyigit et al. (2019) compared eight different artificial neural network models that could quickly and accurately recognize cyberbullying in posts on Turkish social media. The dataset of 3,000 Turkish tweets served as the foundation for comparing these models and different text-mining techniques. The evaluation's results show that, with an F-measure score of 0.91, YSA2 was the model that most accurately predicted cyberbullying content. The machine learning classifiers from earlier studies that they evaluated performed worse when compared to the proposed YSA2.

Ccolltekin (2020) created the first large corpus of Turkish offensive language that contained 36,232 tweets as sample data. The paper focused on the percentage of offensive language (approximately 19%) in the given tweets and then classified them into subcategories based on the contents of those offensive tweets. Using state-of-the-art text classification methods such as SVM, in a 10-fold cross-validation-based experiment, they obtained Precision of 0.786, Recall of 0.76 and F1 score of 0.773 for the Turkish language.

By gathering tweets on immigrants and filtering them using words typically used to describe immigrants, Beyhan et al. (2022) produced the Refugees dataset. As a starting point for their dataset, they created a hate speech detection system using the transformer architecture (BERTurk). When the system was validated using 5-fold cross-validation on the dataset for the Istanbul Convention it achieved accuracy of binary classification is 0.77 and achieved 0.71 on the Refugee dataset. Additionally, for the datasets pertaining to the Istanbul Convention and Refugees, the authors tested a regression model with 0.66 and 0.83 RMSE on a scale.

**Table 14**
The recent studies on low-resource Persian language cyberbullying detection or related topics[Accuracy(A),Precision(P),F1 Score(F1)]

| Reference | Classifier Model | Highest Score | Language for Model | Dataset Size | Different Label | Origin Sources |
|-----------|------------------|---------------|--------------------|--------------|-----------------|----------------|
| Dehghani et al. (2021) | DNN,BERT | 0.977(A) | Persian | 33,338 | Abusive Not Abusive | Twitter |
| Alavi et al. (2021) | MNB,BERT with ANN, CNN, RNN | improvement 0.02(F1) and 0.1(F1) for English and Persian | Persian | 13,280 | NOT,OFF(34.94%) | Tweets, Instagram |

Coban et al. (2023) developed a Turkish language Facebook content-based model to identify online bullying. They used cross-domain evaluation and profiling of cyber-aggressive users. A macro-average F1 score was used to evaluate the model in a 5-fold cross-validation setup. They got a macro-average F1 of 0.928 using BERT on a variety of datasets from various social media domains.

### 3.2.10. Persian Language

The recent studies on cyberbullying detection and related topics in the Persian language is shown in table 14.

In order to find offensive user phrases in Persian tweets, Dehghani et al. (2021) developed a dataset of 33,338 tweets, 10% of which contained abusive language, and 90% of which did not. They used a simple way to increase the probability of abusive content by applying a fixed list of 648 derogatory Persian terms. The score for applying a simple classification method based on dictionary lookup with that list achieved an accuracy of 0.76. When combined with the Bert language model, the deep neural network recognized abusive words with an accuracy rate of 0.977.

By paying more attention to offensive terms than others, since these expressions have a bigger impact on the intended target, it is possible to increase the efficiency of the algorithms for spotting insulting language. In order to achieve, Alavi et al. (2021) developed a novel "Attention Mask" input based on the notion of "offensive score" in order to enhance the detection of offensive content by BERT-based models. They stated that putting this idea into practice has potential, but selecting the right value for each token can be challenging. In order to find improper ratings, they used an automated Multinomial Naive Bayes statistical approach. Their evaluation metric was the F1-macro score, and in the final section, they modified BERT with ANN, CNN, and RNN combinations to investigate the outcomes of using this methodology with different combinations. The results demonstrated that this strategy enhanced all models' performance. The Persian and English languages increases of 0.02 and 0.1, respectively.

### 3.2.11. Indonesian Language

The recent studies on cyberbullying detection and related topics in the Indonesian language is shown in table 15.

Nurrahmi and Nurjanah (2018) identified cyberbullying victim actors through text analysis and user credibility analysis and educated them about the dangers of cyberbullying. They gathered data from Twitter and created a Web-based labeling tool to classify tweets as cyberbullying or non-cyberbullying. There were 301 cyberbullying tweets, 399 non-cyberbullying tweets, 2,053 negative words, and 129 swear words returned by the tool. Then, they used SVM and kNN for classification. According to the results, SVM obtained the highest F1-score, of 0.67. They discovered 257 Normal Users, 45 Harmful Bullying Actors, 53 Bullying Actors, and 6 Potential Bullying Actors after conducting a user credibility analysis.

The purpose of the research by Margono (2019) was to identify cyberbullying indicators in written content, as well as to propose and develop effective analysis models for detecting cyberbullying activities on social networks for the Indonesian language. By developing systematic measurement methods and techniques, this study aimed to propose an effective model for discovering patterns of insulting words, which can further aid in accurately detecting cyberbullying messages. In case of bullying detection, the accuracy are 1, 0.9997 and 0.9999 for Naïve Bayes, Decision Tree, and Neural Network respectively. The analysis model created in this study used association rules and a number of classification techniques (Naïve Bayes, Decision Tree, Neural Network). The methods have proven effective in detecting cyberbullying messages on social media.

**Table 15**
The recent studies on low-resource Indonesian language cyberbullying detection or related[**Accuracy(A),Precision(P),F1 Score(F1)**]

| Reference | Classifier Model | Highest Score | Language for Model | Dataset Size | Different Label | Origin Sources |
|---|---|---|---|---|---|---|
| Nurrahmi and Nurjanah (2018) | SVM , KNN | 0.67(F1) | Indonesian | 700 | Non-cyberbullying, Cyberbullying (43%) | Twitter |
| Margono (2019) | NB, DT, Neural Network | 1(A), 0.9997(A) 0.9999(A) | Indonesian | 152,843 | Cyberbullying, Non-cyberbullying | Twitter |
| Laxmi et al. (2021) | SVM+RF, CNN | 0.6508(F1) | Indonesian | 1,425 | Cyberbullying, Non-cyberbullying | Twitter |

**Table 16**
The recent studies on low-resource Swahili language cyberbullying detection or related topics[**Accuracy(A),Precision(P),F1 Score(F1)**]

| Reference | Classifier Model | Highest Score | Language for Model | Dataset Size | Different Label | Origin Sources |
|---|---|---|---|---|---|---|
| Njovangwa and Justo (2021) | Rule based approach, Character percentage matching techniques | 0.97(F1) | Swahili | 400 | Abusive(50%) Non Abusive | TAFORI forum |

Laxmi et al. (2021) developed a cyberbullying model in Indonesia language by gathering data from Twitter by searching for keywords related to cyberbullying incidents. The data was then combined with previous research findings. They received 1,425 tweets in total, with 393 labeled as cyberbullying and 1,032 labeled as non-cyberbullying and built a Doc2Vec model for feature extraction as well as a classifier model that used the baseline classification method (SVM and RF) and CNN. According to the results, the classifier based on CNN and Doc2vec obtained the highest F1-score, of 0.6508.

### 3.2.12. Swahili Language

The recent studies on cyberbullying detection and related topics in Swahili language is shown in table 16.

The problem of Swahili and English abusive word variations spreading on social media has been addressed by Njovangwa and Justo (2021).To detect abusive terms efficiently, they employed a system that combined character percentage matching with rule-based techniques. The evaluation's results reached F1 score ratio of 0.97 and an accuracy ratio of 0.96, both of which were statistically significant at the 0.5 level.

### 3.2.13. South African Language

The recent studies on cyberbullying detection and related topics in South African language is shown in table 17.

Oriola and Kotze (2020a) mainly tried to work with the low-resource languages of South Africa like Afrikaans, Zulu, and Sesotho with English mix-codes to detect offensive tweets and hate speech. They used multi-tier meta-learning models like Logistic Regression (LogReg), Support Vector Machine (SVM), Random Forest (RF) and Gradient Boosting (GB) algorithms. After completing the evaluation, their research showed that SVM, RF, and GB had the best overall accuracy of 0.671.

**Table 17**

The recent studies on low-resource South African language cyberbullying detection or related topics[Accuracy(A),Precision(P),F1 Score(F1)]

| Reference | Classifier Model | Highest Score | Language for Model | Dataset Size | Different Label | Origin Sources |
|---|---|---|---|---|---|---|
| Oriola and Kotze (2020a) | Logistic Regression, SVM, RF, GB | 0.894(A) | Afrikaans, Zulu, Sesotho with English mix-codes | 7,100, 4,500, 4,102 | Hate(2.29%) Offensive(8.66%) Free Speech | Twitter |
| Oriola and Kotze (2020b) | Logistic Regression, SVM, Neural Networks. | 0.97 (F1) | South African | 10,245 | Abusive(4.99%) Non Abusive | Twitter |

**Table 18**

The recent studies on low-resource Vietnamese language cyberbullying detection or related topics[Accuracy(A),Precision(P),F1 Score(F1)]

| Reference | Classifier Model | Highest Score | Language for Model | Dataset Size | Different Label | Origin Sources |
|---|---|---|---|---|---|---|
| Van Huynh et al. (2019) | BiGRU, LSTM, CNN | 0.70576(F1) | Vietnamese | 25,431 | Clean Offensive(5.02%) Hate(3.49%) | Facebook |
| Do et al. (2019) | SVM, Bi-LTSM, LR, GRU, CNN | 0.7143(F1) | Vietnamese | 25,431 | Clean Offensive(5.02%) Hate(3.49%) | Facebook |
| Luu et al. (2021) | Text CNN, GRU,BERT, XLM-R, DistilBERT | 0.6269(F1) | Vietnamese | 33,400 | Hate(10.52%) Offensive(6.77%) Clean | Facebook |
| Luu et al. (2022) | BERT, PhoBERT | 0.8688(A), 0.9060(A) | Vietnamese | 33400, 10,000 | Hate(10.52%) Offensive(6.77%) Clean | Twitter, Facebook, Instagram, Reddit, Tiktok, VnExpress |

Oriola and Kotze (2020b) focused on detecting abusive low-resourced South African language on Twitter. Using South African English corpora as a dataset, they applied Logistic Regression, Support Vector Machine, and Neural Networks. As a result, they achieved an F1 score of 0.97 for LR and 0.95 for SVM.

### 3.2.14. Vietnamese Language

The recent studies on cyberbullying detection and related topics in the Vietnamese language is shown in table 18. Van Huynh et al. (2019) shows extension of combining various deep neural networks, such as BiGRU, LSTM, and CNN models. They experimented on a dataset in the Vietnamese language, which they explored in a competition setting that comprised several core NLP tasks, like data collection, preprocessing, modeling, and evaluating the performance for classification models trained on three categories, namely clean, offensive and hate.

Do et al. (2019) provided both a pre-labeled dataset and an unlabeled dataset for social media posts and comments. To categorize posts and comments, they pre-processed the data and built several ML models. In this study, they used bidirectional long short-term memory to create a model that predicts the Clean, Offensive, and Hate categories of social media content. They obtained a comparative score of 0.7143 on the public standard test created for the VLSP 2019 [32] using the same technology.

**Table 19**
The recent studies on low-resource Portuguese language cyberbullying detection or related topics[Accuracy(A),Precision(P),F1 Score(F1)]

| Reference | Classifier Model | Highest Score | Language for Model | Dataset Size | Different Label | Origin Sources |
|---|---|---|---|---|---|---|
| Silva and Roman (2020) | SVM, MLP, Logistic Regression Naïve Bayes | 0.8836, 0.8765(A), 0.8519, 0.7601(A) | Portuguese | 5,668 | Hate(21.67%) No Hate | Twitter |
| Leite et al. (2020) | BERT, Multilingual BERT | 0.76(F1) | Brazilian Portuguese | 21,000 | LGBTQ +phobia(0.35%) Insult(2.52%) Xenophobia(0.07%) Misogyny(0.19%) Obscene(3%) Racism(0.03%) Toxic(7.64%) | Twitter |

In order to identify hate speech in texts published on Vietnamese social media, Luu et al. (2021) created a dataset known as the ViHSD dataset. The more than thirty thousand comments in this dataset were divided into three categories by human annotators: CLEAN, OFFENSIVE, or HATE. The Macro F1-score for the model trained on the dataset using the BERT model was 0.6269. Luu et al. (2021) also provided a fast method for annotating data.

Experiments in the detection of hate speech on social media for another low-resourced language, namely, Vietnamese were conducted by Luu et al. (2022). In their paper, they first used transformer-based ML models and then used ensemble models to get better accuracy. For datasets, they used messages from Twitter, Facebook, Instagram, Reddit, and Tiktok. They also used ViHSD (Luu et al., 2021) dataset as well as UIT-ViCTSD (Nguyen et al., 2021) dataset to perform the classification and got the accuracy of o.8688 for BERT (bert-base-multilingual-cased) for ViHSD dataset. On the other hand, on UIT-ViCTSD dataset, they achieved an accuracy of 0.906 for PhoBERT model.

### 3.2.15. Portuguese Language

The recent studies on cyberbullying detection and related topics in Portuguese language is shown in table 19.

Silva and Roman (2020) used some classic Machine Learning Algorithms to automatically identify hate speech on social media in Portuguese. They used different models like SVM, MLP, Logistic Regression and Naïve Bayes on a dataset of 5,668 tweets, with 1,228 tweets classified as hate speech and performed the experiment in a 10-fold cross-validation procedure to find the best F1 score for the given dataset. As a result, the accuracy for the algorithms SVM, LR, MLP and NB were 0.8836, 0.8765, 0.8519, and 0.7601, respectively.

Leite et al. (2020) proposed a new sizable dataset containing twitter posts in Brazilian Portuguese. The dataset, called ToLD-Br (Toxic Language Dataset for Brazilian Portuguese), contained tweets written only in Brazilian Portuguese. Brazilian Portuguese BERT and Multilingual BERT were the two applied pretrained BERT models. M-BERT(transfer) models were able to obtain a macro-F1 score of 0.76.

### 3.2.16. Slovenian Language

The recent studies on cyberbullying detection and related topics in the Slovenian language were shown in Table 20.

Markov et al. (2021) categorized textual content into classes of hate or non-hate speech and conducted an experiment to assess the impact of stylometric and emotion-based features on hate speech detection. With a balanced dataset of 7,500 samples collected from Facebook they were able to achieve close to 0.73 of F1 score using a BERT langauge model.

Evkoski et al. (2021) proposed a model for topic extraction, hate speech detection, and gathering of details on how communities and retweet networks changed over time. They collected a dataset of fifty thousand samples from Twitter, although harmful samples were only in minority. Still, they were able to reach 0.77 of F1 score using a BERT language model.

---

[32] https://vlsp.org.vn/vlsp2019/eval

**Table 20**
The recent studies on low-resource Slovenian language cyberbullying detection or related topics[**Accuracy(A),Precision(P),F1 Score(F1)**]

| Reference | Classifier Model | Highest Score | Language for Model | Dataset Size | Different Label | Origin Sources |
|---|---|---|---|---|---|---|
| Markov et al. (2021) | BERT | 0.729(F1) | Slovene | 7,500 | Hate(50%) Non hate | Facebook |
| Evkoski et al. (2021) | Traditinal ML,mBERT cseBERT | 0.77(F1) | Slovene | 50,000 | Normal Inappropriate(1.1%) Offensive(20%) Violent(0.15%) | Twitter |
| Miok et al. (2022) | BAN,BERT MCD BERT | 0.686(F1) | Slovene | 4,364 | Hate(53.3%) Non hate | Facebook |
| Evkoski et al. (2022) | mBERt cseBERT | 0.69(F1) | Slovene | 50,000 | Normal Inappropriate(1.1%) Offensive(20%) Violent(0.15%) | Twitter |

Miok et al. (2022) presented a method for classifying hate speech and quantifying the uncertainty of the predictions of the BERT and attention network models as well as Bayesian attention networks (BANs) and Monte Carlo Dropout(MCD) enhanced BERT models. They demonstrated improvement in calibration and prediction performance, yet still reaching below 0.7 of F1 score.

Evkoski et al. (2022) conducted a challenging task of identifying the most significant Twitter accounts that promote hate speech. The authors used BERT-based models to detect hate speech on a level close to 0.7 of F1 score.

### 3.3. Discussion on quality of research on cyberbullying detection in low-resource languages

In this study, we have discussed in detail over sixty pieces of cyberbullying/abusive language/hate speech detection research done in various low-resource languages (see subsection 3.2). Unfortunately, some of those studies were not of high quality, which we perceive as one of the reasons for low-resource language-based research retaining a niche position. Therefore in this section, we highlight the most frequent quality issues in the research we reviewed, with the hope that it will serve as a guide of what to avoid, for the authors of future studies in the specific area of **automatic cyberbullying detection in low-resource languages**, or the general area of **natural language processing in low-resource languages**.

In the case of cyberbullying in the Bangla language (Eshan and Hasan, 2017; Akhter et al., 2018; Hussain et al., 2018; Jahan et al., 2019; Ishmam and Sharmin, 2019; Karim et al., 2021; Sazzed, 2021), the text for the dataset, such as users' public status, and comments were taken from platforms like Facebook, Twitter, YouTube, and other social media platforms. Unfortunately, it is unclear who annotated the data. Moreover, all studies used small or very small data sets (one thousand to four thousand samples), which is roughly ten-fold smaller than high quality datasets (ten thousand to thirty thousand samples) (Romim et al., 2021). Additionally, none of the studies attempted to benchmark their dataset for quality. Due to the fact that mostly classic machine learning techniques (SVM, Naive Bayes, etc.) have been used in many of those papers, the detection accuracy is usually relatively low, or unreliable. Another major problem is the fact that none of the papers have taken into account ethical considerations of data collection (such as preserving user privacy by anonymizing the dataset, etc.).

The standard for highly influential papers in natural language processing, including those focused on abusive language and hate-speech, usually use novel machine learning techniques, such as transformer-based language models (Devlin et al., 2019; Clark et al., 2020; Romim et al., 2021; Aurpa et al., 2022), or hybrid systems (Urbaniak et al., 2022a,b) and conducting the research on large datasets has shown a successful strategy to produce desirable outcomes. These studies also rely on the strong writing skills of the authors and/or professional proofreading. Finally, making the additional effort to have their research published in highly competitive scientific publication venues also adds to the visibility of the work.

In the case of cyberbullying detection research done in Hindi, Dravidian and Marathi language (Pawar and Raje, 2019; Sreelakshmi et al., 2020; Saha et al., 2021; Maity and Saha, 2021; Shibly et al., 2022; Sharma et al., 2022; Madhu et al., 2023; Chakravarthi et al., 2023; Rajalakshmi et al., 2023), many authors have completed research on cyberbullying detection with relatively large datasets (Mandl et al., 2019). Most of those studies followed good research practices which resulted in obtaining satisfying results. On the other hand, other studies (Gaikwad et al., 2021; Glazkova et al., 2021; Mehendale et al., 2022) performed research with limited datasets, did not report how the data annotation was conducted or report other information necessary in terms of the reproducibility of the research.

Apart from the research by Rizwan et al. (2020), most research papers on cyberbullying in Urdu and Roman Urdu (Akhter et al., 2020; Dewani et al., 2021; Akhter et al., 2022; Dewani et al., 2023; Saeed et al., 2023; Akram et al., 2023) worked with a sufficient amount of data, provided enough information required for reproducing the research, and published their research work in reputable venues. However, some publications (Talpur et al., 2020; Das et al., 2021) failed to provide the necessary information (e.g. annotation procedure, model architecture, etc.), or discuss ethical concerns.

Cyberbullying detection has been well reported in Arabic/Arabic dialect by several authors (Haddad et al., 2019; Mouheb et al., 2019; Alshalan and Al-Khalifa, 2020; Alsafari et al., 2020; Boucherit and Abainia, 2022; Alduailaj and Belghith, 2023; Elzayady et al., 2023; Fkih et al., 2023), even employing benchmark datasets in comparison to publicly available datasets[33]. However, some research publications (Faris et al., 2020; Nayel, 2020; AlHarbi et al., 2020; Farid and El-Tazi, 2020) failed to identify data sources, even when comparing their findings to studies done for highly resourced languages (Sazzed, 2021).

In case of Turkish (Ozel et al., 2017; Bozyigit et al., 2019; Ccolltekin, 2020), Persian (Dehghani et al., 2021; Alavi et al., 2021; Coban et al., 2023), Indonesian (Margono, 2019), Vietnamese (Luu et al., 2021, 2022) and Portuguese (Leite et al., 2020) language oriented research, the majority of studies used adequate data, gave the necessary details for replication, and were published in reputed venues. In contrast, several papers (Nurrahmi and Nurjanah, 2018; Van Huynh et al., 2019; Do et al., 2019; Silva and Roman, 2020; Laxmi et al., 2021; Beyhan et al., 2022) neglected to highlight ethical issues or offer the relevant details (such as the annotation technique or model architecture).

When it comes to cyberbullying detection in languages of the African Continental group, it is quite encouraging that the majority of the low-resource languages and dialects of the African Continental group appeared at least once in our survey (Oriola and Kotze, 2020b,a; Njovangwa and Justo, 2021). However, some studies used smaller datasets, or did not report with sufficient details when compared to others Haddad et al. (2019).

From the above analysis, we were able to produce a list of the most prevalent issues found in papers on cyberbullying detection in low-resource languages, which includes the following:

- poor definition of the problem,

- small-sized dataset,

- unknown dataset source,

- paper poorly written,

- paper too short (not enough information),

- limited experiments,

- superficial or nonexistent field survey,

- journal publisher has a low range of dissemination.

Based on the issues listed above, we generalized several most prevalent research gaps, and proposed a list of good practices to follow, when doing research on automatic cyberbullying detection in low-resource languages.

---

[33]https://hatespeechdata.com/

**Table 21**
Linguistic similarity metric between all surveyed languages [**Afrikaans (AF), Arabic Egyptian (AE), Arabic (AR), Arabic Tunisian (AT), Bangla (BA), Hindi (HI), Indonesian (IN), Kannada (KA), Malayalam (ML), Marathi (MA), Nepali (NE), Persian (PE), Portuguese (PO), Sesotho (SE), Sinhala (SI), Slovenian (SL), Swahili (SW), Tamil (TA), Turkish (TU), Urdu (UR), Vietnamese (VI), and Zulu (ZU)**]. No score means lack of sufficient data to calculate the score.

| | AF | AE | AR | AT | BA | HI | IN | KA | ML | MA | NE | PE | PO | SE | SI | SL | SW | TA | TU | UR | VI | ZU | |
|---|---|---|---|---|---|---|---|---|---|---|---|---|---|---|---|---|---|---|---|---|---|---|---|
| AF | 0.00 | 0.25 | | 0.25 | 0.16 | 0.26 | 0.15 | 0.16 | 0.25 | 0.12 | 0.19 | 0.19 | 0.20 | 0.21 | 0.17 | 0.23 | 0.22 | 0.19 | 0.25 | 0.11 | 0.22 | 0.31 | AF |
| AE | 0.25 | 0.00 | 0.18 | 0.07 | 0.25 | 0.20 | 0.23 | 0.22 | 0.25 | 0.28 | 0.29 | 0.18 | 0.18 | 0.19 | 0.26 | 0.25 | 0.17 | 0.23 | 0.24 | 0.25 | 0.22 | 0.19 | AE |
| AR | | 0.18 | 0.00 | | 0.28 | 0.16 | 0.14 | 0.30 | 0.28 | 0.26 | 0.27 | 0.21 | 0.12 | 0.16 | 0.39 | 0.08 | 0.14 | 0.31 | 0.27 | 0.20 | 0.13 | 0.24 | AR |
| AT | 0.25 | 0.07 | | 0.00 | 0.26 | 0.20 | 0.34 | 0.37 | | | | 0.11 | | 0.16 | | 0.28 | 0.20 | 0.31 | 0.23 | 0.30 | 0.30 | 0.23 | AT |
| BA | 0.16 | 0.25 | 0.28 | 0.26 | 0.00 | 0.16 | 0.22 | 0.08 | 0.17 | 0.03 | 0.12 | 0.20 | 0.33 | 0.27 | 0.09 | 0.19 | 0.22 | 0.05 | 0.13 | 0.30 | 0.23 | 0.26 | BA |
| HI | 0.26 | 0.20 | 0.16 | 0.20 | 0.16 | 0.00 | 0.25 | 0.16 | 0.14 | 0.14 | 0.16 | 0.18 | 0.20 | 0.24 | 0.15 | 0.12 | 0.21 | 0.16 | 0.16 | 0.02 | 0.23 | 0.24 | HI |
| IN | 0.15 | 0.23 | 0.14 | 0.34 | 0.22 | 0.25 | 0.00 | 0.26 | 0.31 | 0.30 | 0.27 | 0.21 | 0.21 | 0.18 | 0.21 | 0.19 | 0.20 | 0.32 | 0.30 | 0.18 | 0.10 | 0.26 | IN |
| KA | 0.16 | 0.22 | 0.30 | 0.37 | 0.08 | 0.16 | 0.26 | 0.00 | 0.14 | 0.13 | 0.07 | 0.21 | 0.28 | 0.30 | 0.12 | 0.24 | 0.21 | 0.09 | 0.13 | 0.19 | 0.27 | 0.21 | KA |
| ML | 0.25 | 0.25 | 0.28 | | 0.17 | 0.14 | 0.31 | 0.14 | 0.00 | 0.08 | 0.05 | 0.21 | 0.28 | 0.27 | 0.08 | 0.25 | 0.24 | 0.09 | 0.11 | 0.22 | 0.24 | 0.27 | ML |
| MA | 0.12 | 0.28 | 0.26 | | 0.03 | 0.14 | 0.30 | 0.13 | 0.08 | 0.00 | 0.06 | 0.25 | 0.31 | 0.33 | 0.09 | 0.30 | 0.25 | 0.11 | 0.14 | 0.18 | 0.28 | 0.28 | MA |
| NE | 0.19 | 0.29 | 0.27 | | 0.12 | 0.16 | 0.27 | 0.07 | 0.05 | 0.06 | 0.00 | 0.24 | 0.30 | 0.30 | 0.12 | 0.27 | 0.23 | 0.07 | 0.09 | 0.19 | 0.29 | 0.25 | NE |
| PE | 0.19 | 0.18 | 0.21 | 0.11 | 0.20 | 0.18 | 0.21 | 0.21 | 0.21 | 0.25 | 0.24 | 0.00 | 0.22 | 0.20 | 0.22 | 0.20 | 0.19 | 0.22 | 0.16 | 0.14 | 0.22 | 0.22 | PE |
| PO | 0.20 | 0.18 | 0.12 | | 0.33 | 0.20 | 0.21 | 0.28 | 0.28 | 0.31 | 0.30 | 0.22 | 0.00 | 0.19 | 0.18 | 0.09 | 0.21 | 0.32 | 0.26 | 0.19 | 0.16 | 0.29 | PO |
| SE | 0.21 | 0.19 | 0.16 | 0.16 | 0.27 | 0.24 | 0.18 | 0.30 | 0.27 | 0.33 | 0.30 | 0.20 | 0.19 | 0.00 | 0.18 | 0.17 | 0.15 | 0.27 | 0.29 | 0.30 | 0.18 | 0.16 | SE |
| SI | 0.17 | 0.26 | 0.39 | | 0.09 | 0.15 | 0.21 | 0.12 | 0.08 | 0.09 | 0.12 | 0.22 | 0.18 | 0.18 | 0.00 | 0.22 | 0.27 | 0.06 | 0.12 | 0.22 | 0.26 | 0.29 | SI |
| SL | 0.23 | 0.25 | 0.08 | 0.28 | 0.19 | 0.12 | 0.19 | 0.24 | 0.25 | 0.30 | 0.27 | 0.20 | 0.09 | 0.17 | 0.22 | 0.00 | 0.22 | 0.26 | 0.15 | 0.17 | 0.15 | 0.30 | SL |
| SW | 0.22 | 0.17 | 0.14 | 0.20 | 0.22 | 0.21 | 0.20 | 0.21 | 0.24 | 0.25 | 0.23 | 0.19 | 0.21 | 0.15 | 0.27 | 0.22 | 0.00 | 0.23 | 0.23 | 0.22 | 0.24 | 0.08 | SW |
| TA | 0.19 | 0.23 | 0.31 | 0.31 | 0.05 | 0.16 | 0.32 | 0.09 | 0.09 | 0.11 | 0.07 | 0.22 | 0.32 | 0.27 | 0.06 | 0.26 | 0.23 | 0.00 | 0.11 | 0.23 | 0.30 | 0.25 | TA |
| TU | 0.25 | 0.24 | 0.27 | 0.23 | 0.13 | 0.16 | 0.30 | 0.13 | 0.11 | 0.14 | 0.09 | 0.16 | 0.26 | 0.29 | 0.12 | 0.15 | 0.23 | 0.11 | 0.00 | 0.23 | 0.28 | 0.23 | TU |
| UR | 0.11 | 0.25 | 0.20 | 0.30 | 0.30 | 0.02 | 0.18 | 0.19 | 0.22 | 0.18 | 0.19 | 0.14 | 0.19 | 0.30 | 0.22 | 0.17 | 0.22 | 0.23 | 0.23 | 0.00 | 0.20 | 0.30 | UR |
| VI | 0.22 | 0.22 | 0.13 | 0.30 | 0.23 | 0.23 | 0.10 | 0.27 | 0.24 | 0.28 | 0.29 | 0.22 | 0.16 | 0.18 | 0.26 | 0.15 | 0.24 | 0.30 | 0.28 | 0.20 | 0.00 | 0.29 | VI |
| ZU | 0.31 | 0.19 | 0.24 | 0.23 | 0.26 | 0.24 | 0.26 | 0.21 | 0.27 | 0.28 | 0.25 | 0.22 | 0.29 | 0.16 | 0.29 | 0.30 | 0.08 | 0.25 | 0.23 | 0.30 | 0.29 | 0.00 | ZU |
| | AF | AE | AR | AT | BA | HI | IN | KA | ML | MA | NE | PE | PO | SE | SI | SL | SW | TA | TU | UR | VI | ZU | |

### 3.3.1. Comparative reflections between different languages

Within the over 70 papers reviewed in this study, we covered overall 23 different languages of different countries (See Figure 6). These languages differ from one another in terms of semantic, morphological, and syntactic structure. Therefore a discussion on how cyberbullying is expressed in each of those languages needs to be performed in accordance with how close, or similar, those languages are to one another. To draw linguistic similarity based on these terms, previous study proposed a language similarity metric qWALS< or quantified World Atlas of Language Structures (Eronen et al., 2023). Based on that study, we calculated distances among all among languages covered in this survey. In the light of this study, it can be seen how similar in general the covered languages are. In this metric two languages are more linguistically similar to one another when their score is lower (score 0.00 indicates a 100% linguistic similarity between the two languages), and more distant when this score is higher. For instance, Malayalam and Nepali have a score of 0.05, which indicates a higher degree of linguistic similarity (or - small distance) between the two languages. Linguistic similarity metric between all surveyed languages can be shown in Table 21.

Apart from linguistic similarity, the use of bullying-related language to settle disputes is widespread in language communities with low resources. For instance, using the word বাল (Hair), including insults, to make a point, may be acceptable in some communities in particular areas of West Bengal[34]. However, this kind of conduct would be regarded as offensive in Bangladesh[35] cultures. For instance, in Bangladesh, the word বাল is typically used in a vulgar meaning (pubic hair).

Additionally, we noticed the following reflections among the cyberbullying detection studies done between these languages.

1. Machine learning models: The following low-resource languages perform satisfactory in terms of machine learning models: Bangla (Eshan and Hasan, 2017; Akhter et al., 2018; Awal et al., 2018; Islam et al., 2021), Hindi (Tarwani et al., 2019; Pawar and Raje, 2019; Sreelakshmi et al., 2020; Mehendale et al., 2022), Sinhala (Sandaruwan et al., 2019; Amali and Jayalal, 2020), Dravidian (Shibly et al., 2022; Subramanian et al., 2022), Zulu (Oriola and Kotze, 2020a), Sesotho (Oriola and Kotze, 2020a), Portuguese (Silva and Roman, 2020), Arabic (Mouheb et al., 2019; AlHarbi et al., 2019; Alduailaj and Belghith, 2023; Fkih et al., 2023), and Urdu (Akhter et al., 2020; Dewani et al., 2023). On the other hand, machine learning model training was less successful for languages such as Napali (Niraula et al., 2021), Vietnamese (Do et al., 2019), Turkish (Ccolltekin, 2020), Indonesian (Nurrahmi and Nurjanah, 2018; Laxmi et al., 2021), and Slovenian (Evkoski et al., 2021).

---

[34]https://en.wikipedia.org/wiki/West_Bengal
[35]https://en.wikipedia.org/wiki/Bangladesh

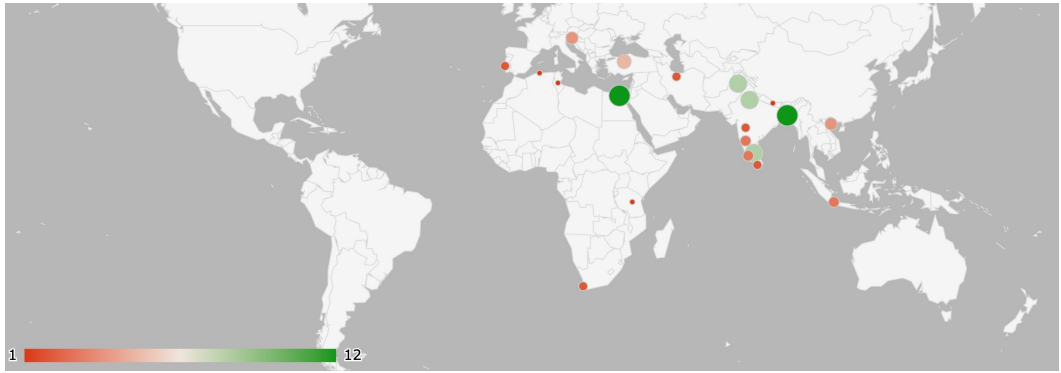

**Figure 6:** Areas of languages surveyed

2. Deep learning models: In terms of deep learning models, previous research managed to develop satisfactory-performing models for such low-resource languages, as Indonesian (Margono, 2019), Persian (Dehghani et al., 2021; Alavi et al., 2021), Bangla (Karim et al., 2021; Aurpa et al., 2022), Hindi (Maity and Saha, 2021; Anand et al., 2023; Madhu et al., 2023), Dravidian (Sharif et al., 2021; Chakravarthi et al., 2023), South African (Oriola and Kotze, 2020b), Arabic (Alsafari et al., 2020; Alshalan and Al-Khalifa, 2020), and Urdu (Saeed et al., 2023; Akram et al., 2023), while for languages such as Algerian dialect (Boucherit and Abainia, 2022), and Vietnamese (Van Huynh et al., 2019) it was less effective.

3. Multilingual models: By utilizing the shared features of related languages, multilingual langaeuege models enhance the performance of NLP tasks for such low-resource languages such as Bangla (Karim et al., 2021; Aurpa et al., 2022), Hindi (Sharma et al., 2022), Dravidian (Saha et al., 2021; Subramanian et al., 2022; Rajalakshmi et al., 2023), Marathi (Glazkova et al., 2021), Urdu (Saeed et al., 2023; Akram et al., 2023), Persian (Dehghani et al., 2021; Alavi et al., 2021), Vietnamese (Luu et al., 2022). However, the pronounced phonological and grammatical differences might be problematic for multilingual language models when it comes to the languages like Arabic (Alshalan and Al-Khalifa, 2020), Turkish (Beyhan et al., 2022) and Slovenian (Markov et al., 2021; Evkoski et al., 2021; Miok et al., 2022; Evkoski et al., 2022).

### 3.4. Most prevalent research gaps and proposal of good practices

Apart from the critical analysis of the general quality of research in the area of focus, we looked at the research gaps that have been prevalent even in research of satisfying quality. Moreover, based on those research gaps, as well as on quality issues discussed in section 3.3, we propose a set of good practices to follow in our research, as well as all future research in the area of automatic cyberbullying detection for low-resource languages.

Among the numerous studies described in subsection 3.2, some work has been done on the official languages of various countries (Beyhan et al., 2022; Luu et al., 2022; Aurpa et al., 2022), even though with extremely limited datasets, and without considering ethical issues. However, little work has been done on the various dialects used in the country (Farid and El-Tazi, 2020). Thus, the first main research gap is that **very little work is done in the dialects of low-resource languages**. This in fact extends to high-research languages, as dialects of those high-resource languages can be considered low-resource languages themselves.

Some good practices for the research on cyberbullying detection in general, have been laid out by Ptaszynski (2022) (see also Figure 7). These include the following:

- Data should be collected to the furthest extent possible, however, data collected completely randomly and without sufficient post-processing will result in a large number of low-quality samples (mixture of languages, extraction errors, noise, etc.).

- The careful attention paid during the process of data collection should extend to preserving the balance of harmful to non-harmful samples in the dataset (with the ideal being 50% of harmful and 50% non-harmful).

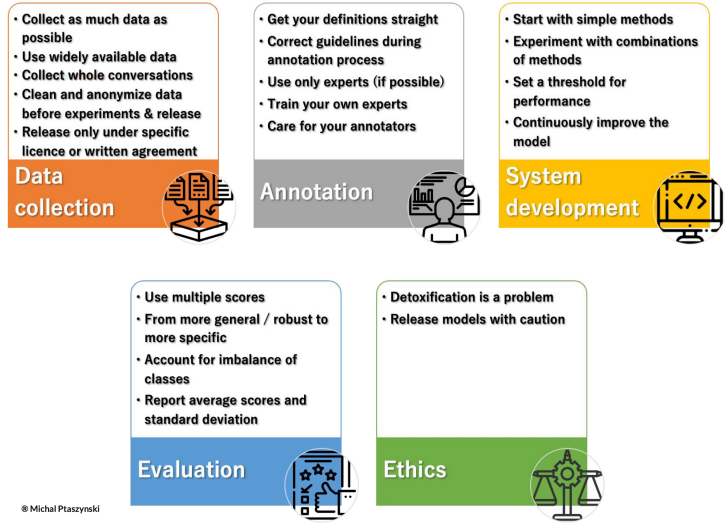

**Figure 7**: List of good practices in automatic cyberbullying detection (Ptaszynski, 2022).

- The process of data collection should be done in cooperation with experts, such as cyberbullying experts, Internet patrol group members, Parent-Teacher Association members, or even the police, if possible.

- The annotation process should be based on official well-thought-out definitions of the collected phenomenon, with clearly understandable taxonomy and annotation guidelines supported with examples.

- The creation of the classifier or analysis tool, should start with the simplest methods possible, even simple rule-based methods, followed by classic machine learning methods (SVM, Random Forest, Naïve Bayes, etc.), including combinations and ensembles of those methods, and only when those methods fail to reach satisfying results, more processing-heavy methods should be applied, such as deep learning, including word embeddings, or transformer models.

- Models in production require continuous improvement due to the constant changes in the language used on the Internet.

- Regarding the evaluation procedures, reporting averages of multiple experiment runs with standard deviations will allow for finding

    a  performance boundaries of the model,
    b  how stable the model is,
    c  best random seed.

- Ethical considerations of creating such models should be taken into account (for a longer discussion on ethics, see subsection 5.1).

The above good practices would allow for a good baseline in research conduct for future cyberbullying detection research, especially considering low-resource languages.

## 4. Initial attempt at creating and testing a dataset for cyberbullying detection in Chittagonian dialect of Bangla language

By studying earlier research publications on the topic of cyberbullying detection and adhering to good practices, we decided to provide an additional contribution by creating and initially benchmarking a dataset for identifying cyberbullying in the Chittagonian dialect of Bangla language[36]. Chittagonian is an Indo-Aryan language[37], and a dialect of Bangla spoken in many places in Bangladesh. However, although Chittagonian dialect speakers may be familiar with Bengali culture and language, both are incomprehensible to one another.

**Table 22**
Statistics: sources of data.

| Social Media Platform | Collected Sample |
|---|---|
| Facebook | 2,500 |
| YouTube | 644 |
| TikTok | 618 |
| News Portal | 567 |
| Personal Blogs | 356 |
| WhatsApp | 335 |
| Imo | 80 |
| **Total=** | **5,100** |

In this section, we discuss several steps taken during the collection and validation of this dataset. We started our task by defining the standard or rules set by various social media platforms and organizations to define cyberbullying and other harmful types of language. As social media platforms become mainstream means for people to acquire and exchange information, cyberbullying becomes more prevalent by the day. Therefore, automatically detecting and filtering cyberbullying entries is an important and urgent matter, especially for those areas of the Internet where low-resource languages, such as Chittagonian, are used.

## 4.1. Cyberbullying Detection in Chittagonian dialect of Bangla (CBDCB)

The next subsections go through the annotation guidelines, annotation procedure, and quantitative and qualitative findings we made with the dataset.

### 4.1.1. Data collection

As there was no available raw dataset of Chittagonian, one of the first major challenges in this study was collecting the data. We collected a total 5,100 comments/posts in the Bangladeshi Chittagonian dialect manually from various social media between April 2 and May 1, 2022(see Table 22).The comments and post included have covered a variety of topics, including politics, religion,COVID-19 coronavirus circumstance, sports, and more. Moreover, the dataset was collected mostly from various Facebook groups, pages, YouTube, blogs, and news portals like Chatgaiya Express [38], and Chatgaiya Tourist Gang[39].

In the final version of the dataset we included the original text alongside the labels 1/Yes and 0/No for Bullying/harmful and Non-bullying/non-harmful, respectively. Table 23 shows five examples from our dataset.

### 4.1.2. Standard used in data annotation

As the automatic detection of cyberbullying is an urgent matter, there is a need for a standardized taxonomy of the categories related to the phenomenon. The problem has been defined differently in the literature, but some fundamental elements remain the same. Beginning with laypeople-focused definitions, Cambridge Dictionary defines **cyberbullying** as an activity *utilizing the Internet to harm or threaten another person*[40]; the Oxford Dictionary defines it as the use of *messages on social media, emails, text messages, etc. to frighten or upset somebody*[41]; Medical Dictionary Online defines it as an aggressive behavior, physical or verbal, intended to cause harm or distress[42]. Apart from the above brief definitions, the Facebook community established ground rules for what types of content are not permitted, such as posts containing cyberbullying, violence, menace, reprimand, or personification[43]. A community can set up a policy for dealing with cyberbullying content, which can be based on expressing hatred or discrimination

---

[36]https://en.wikipedia.org/wiki/Chittagonian_language
[37]https://en.wikipedia.org/wiki/Indo-Aryan_languages
[38]https://web.facebook.com/groups/1535657099839469
[39]https://web.facebook.com/groups/999621230477551
[40]https://dictionary.cambridge.org/dictionary/english/cyberbullying
[41]https://www.oxfordlearnersdictionaries.com/definition/english/cyberbullying
[42]https://www.online-medical-dictionary.org/definitions-b/bullying.html
[43]https://www.facebook.com/help/216782648341460

---

**Table 23**
Five examples from the Dataset.

| Data | English Translation | Source | Type | Target Category | Label |
|---|---|---|---|---|---|
| তোরে কুত্তাইও বিয়া গইরতু নু। | A dog will not marry you. | Facebook | Comment | Women/Girls | 1 (harmful) |
| আরা চট্টগ্রামের মানুষ সেরা | We people of Chittagong are the best | Facebook | Post | Community | 0 (non-harmful) |
| বাংলাদেশত নতুন নতুন বেইশ্যা দেহা যার। | New prostitutes are appearing in Bangladesh. | YouTube | Comment | Women/Girls | 1 (harmful) |
| শুভ জন্মদিন প্রিয় গ্রুপ,,, বউত বউত | "Happy Birthday dear Group,,, Many Many" | TikTok | Post | Group | 0 (non-harmful) |
| তুত্তা | You are a bitch | News Portal | Comment | Men/Boys | 1 (harmful) |

against such user attributes like age (ageism), race (racism), disability, ethnicity, immigration status, religion, sex, sexual orientation, or gender identity[44]. Twitter considers abusive behavior to be an attempt to infiltrate, threaten, or silence other people[45]. Recently there have been a number of initiatives to organize shared tasks for offensive language detection, also for low-resource languages (Kumaresan et al., 2021), which also propose their definitions and taxonomies of cyberbullying. For example, Ccolltekin (2020) defines cyberbullying as the use of language that would degrade the dignity of a person and includes profanity and racial, ethnic, or sexist insults, as well as violence and threats, or making statements that are meant to be derogatory or embarrassing to someone based on their real or perceived race, color, religion, gender, sexual orientation, or gender identity. Based on the above definitions, we propose a taxonomy for the need of creating a dataset of cyberbullying posts and comments in the Chittagonian dialect. We explain this taxonomy in the form of annotation guidelines, below.

### 4.1.3. Data annotation guidelines

Below we present a set of definitions for categories of cyberbullying language we collected for the proposed CBDCB dataset. In summary, the language of bullying is aimed at spreading negativity in the online community. Different studies also expand the cyberbullying category to include other general categories of abusive or offensive language, like "hate speech," "inciteful language," "religious hatred," "political hatred," and "communal hatred" (hatred against specific community) (Ishham and Sharmin, 2019). On the other hand, Romim et al. (2021) specified seven topical areas related to cyberbullying (sports, entertainment, crime, religion, politics, celebrity, memes).

After reviewing previous work on cyberbullying detection in low-resource languages, their definitions, and datasets, we came to the consensus that a sentence is classifiable as cyberbullying if it falls into any of the five categories listed below.

**Targeting individuals:** Comments or posts with threats, offensive language, bullying, obscenity, vulgarity or swearing towards an individual. Below is an example of such an entry from our dataset.
Example: সানা ওজ্ঞা খানকি মাইয়াপোয়া English: Sana[46] is a very bad girl)

**Targeting a larger group:** Comments or posts containing threat, offensive language, bullying, obscenity, vulgarities or swearing which is aimed at a group or an organization.
Example: ডিভি কুত্তাচোদা পার্টি (English: DV[47] is a dog fucking party)

**Bullying targeted others without mentioning any identification:** Comments or post having, threat, offence, bully, obscenity, vulgar or swearing which aiming towards except the above mentioned two.
Example: খানকির ফুয়ার বাঁচি থাহনর দরহার নাই (English: Whore's son does not need to live)

**Untargeted bullying and vulgar words:** Comments or posts having, threat, offence, bully, obscenity, swearing or vulgar words but not targeting any one.
Example: এহন মেনসর জিবন ইয়েনকুত্তার তুন হারাপ (English: Now human's life is worse than fox and dog)

---

[44]https://support.google.com/youtube/answer/2802268?hl=en#
[45]https://help.twitter.com/en/rules-and-policies/hateful-conduct-policy
[46]Name changed due to personal information
[47]Name changed due to political group information

**Table 24**
Summary of annotations.

| Category | Number of Comments |
|---|---|
| Bullying | 2610 |
| Not Bullying | 2444 |
| Ignored | 25 |
| Conflicts | 21 |

**Table 25**
Dataset statistics.

| Attributes | Bullying | Not Bullying |
|---|---|---|
| Number of Words | 22346 | 36079 |
| Vocabulary size | 3324 | 5236 |
| Highest number of words in a sentence | 19 | 60 |
| Lowest number of words in a sentence | 2 | 10 |
| Average number of words in a sentence | 8 | 20 |

**Bullying sentence without having any slang or vulgar words:** Sentence spread negativity without having any types of slang and vulgar words.

Example: **** মন্ত্রী দুর্নীতির টে দিয় কানাডাত সম্পদ বানায়ে। (English: ****[48] minister has made wealth in Canada with corruption money)

In the annotation process, if an entry was entirely written in another language (e.g., English, or another low-resource language, like Chakma[49], Tripura[50] or Marma[51] etc.), or if it was ambiguous, or difficult to understand, it was ignored.

### 4.1.4. Data annotation results

The dataset was annotated by three annotators, one male, and two female annotators. To ensure the success of the data annotation process, we trained the annotators in accordance with the data annotation standard and guidelines outlined in subsections 4.1.2 and 4.1.3. All annotators were Chittagonian native speakers with higher education degrees (Master's or Bachelor's degree). They were told to annotate each paragraph with one of two options (Bullying or Not Bullying).

We asked the three annotators to classify 5,100 collected comments as Bullying, or Not Bullying, and for the entire dataset, we received 15,300 judgments.

All three annotators agreed on most of the entries, 25 comments were ignored(discarded from the dataset after discussion with annotators), 2,610 comments were annotated as Bullying, 2,444 comments were annotated as Not Bullying. The annotators disagreed in 21 cases(discarded from the dataset after discussion with annotators) (see Table 24).

One of the final goals of this study was to create an automated system for detecting cyberbullying in the Chittagonian dialect irrespective of its type. Therefore why we categorized the comments under two general categories: Bullying and Not Bullying. Table 25 represents the statistics of our dataset.

### 4.1.5. Quality evaluation of data and annotations

Before validation, we verified the statistical reliability of the dataset and the annotations. Specifically, we tested the reliability of annotations by verifying the agreements between the annotators. To do that we used two measures: pairwise percentage agreement (PPA) (Artstein and Poesio, 2008) and inter-annotator agreement (Cohen's kappa) (Cohen, 1968). The agreements between all pairs of annotators were very strong, reaching over 95% of PPA.

---

[48] Name not mentioned due to political personal information
[49] https://en.wikipedia.org/wiki/Chakma_language
[50] https://en.wikipedia.org/wiki/Languages_of_Tripura
[51] https://en.wikipedia.org/wiki/Arakanese_language

**Table 26**
Numbers of samples annotators agreed upon (pairwise).

| Category | Bullying | Not Bullying |
|---|---|---|
| 1st Annotator Vs 2nd Annotator | 2510 | 2544 |
| 1st Annotator Vs 3rd Annotator | 2720 | 2334 |
| 2nd Annotator Vs 3rd Annotator | 2600 | 2454 |

**Table 27**
inter-rater reliability results

| Category | Pairwise Percentage Agreement | Cohen's kappa |
|---|---|---|
| 1st Annotator Vs 2nd Annotator | 96.14% | 0.928 |
| 1st Annotator Vs 3rd Annotator | 95.80% | 0.946 |
| 2nd Annotator Vs 3rd Annotator | 96.03% | 0.948 |

To confirm the result obtained by PPA, we used another measure, namely, Cohen's kappa (Cohen, 1968), which takes into account the probability of agreement by chance. It is a correlation coefficient between annotators that varies from 0 to +1, with 0 representing no agreement and 1 representing complete agreement[52].

Cohen's kappa values for the two categories (Bullying and Not Bullying) were also very high, reaching over 0.9 for all pairs of annotators, indicating very strong agreement among annotators (see Tables 26 and 27).

We manually verified the annotators' interpretation of the data on the 25 comments that were ignored. All included comments written in a language different than Chittagonian.

Moreover, we also looked at the 21 comments that had conflicts between three annotators. Because we chose three persons from different backgrounds, we observed that they did not agree on the topics related to caste and religion. This could be considered as an annotator's bias.

In the end, the dataset comprised of 2,610 Bullying samples out of all collected 5,100 comments, indicating that the dataset is well balanced.

## 4.2. Data preprocessing

There is a number of techniques for preprocessing data to make it processable by machine learning classifiers. Most common of those techniques include n-grams, TF-IDF, or word2Vec, etc., and are used to convert the text data into numeric, or vector form, or – vectorization. However, before vectorization, it is necessary to clean and prepare the data for further processing, or – to perform data preprocessing.

Data preprocessing is the process of cleaning text data and preparing it to fit a machine-readable form. Text data, apart from elements that can be useful in classification, such as words, and their context (surrounding words), or punctuation (e.g., exclamation marks, like "!" can inform statistical and rule-based models of emotional load expressed in sentences Ptaszynski et al. (2017), comprises also of noise of different kinds, such as redundant punctuation, emoji, emoticons, special characters, etc. Below, we describe step by step, all the preprocessing techniques applied in this study (see also Figure 28 for a visualization of how a sentence is manipulated at each step of preprocessing).

**Removing punctuations:** Text data contains a large variety of punctuation and special characters, which often does not have much impact the meaning of a sentence. Previous studies showed that the removal of punctuation can improve text classification, including automatic cyberbullying detection, when classic ML algorithms are applied in classification (Eronen et al., 2021). Therefore in our study, we also removed all punctuation from the analyzed text. A list representing all types of punctuation present in our data includes the following: ' !#()*+,-./:;<=>?@[]'|

**Removing emoji and emoticons:** Users have used various types of emojis and emoticons to express their own feelings and emotions in Internet messages. There are two ways to handle this kind of information, namely, to either eradicate them from the text or to replace them with the corresponding text. Although emojis are known to sometimes help with text classification (Li et al., 2020), since in this research we aimed mostly at testing the baseline performance of simple ML classifiers on our dataset, we decided to not take emojis and emoticons into consideration in classification.

---

[52]Cohen's kappa can in some situations reach negative values, but then it is treated as "0". See for example Ptaszynski et al. (2021)

**Removing English characters:** Although Bangladesh is a monolingual country, people sometimes mix in English also while expressing their thoughts. As in our task we focused only on the Chittagonian dialect we deleted all types of characters except those in the Chittagonian dialect.

**Removing Bangla and English digits:**

After careful inspection of dataset samples, we noticed that many cases contained digits and numbers that did not specifically impact meaning. Typically, for the case of digits and numbers, one would use a Named Entity Recognition (NER) tool to specify whether they represent phone numbers, percentages, currencies, etc. However, at present, there exists no NER tool for Chittagonian. Therefore, for the needs of the initial experiment, we deleted the digits and numbers. We also plan to develop a tool for NER in Chittagonian in the future, which will allow more effective processing of such cases.

**Removing stopwords:** Every natural language contains a number of words that do not influence the meaning of a sentence significantly. These are known as stopwords. A typical strategy in such cases is to remove the stopwords to reduce the dimensionality of the dataset Eronen et al. (2021). Some of the examples of stopwords in English include: "a", "the", "is", "are". Some of the common stopwords in Chittagonian languages include 'কি(What)','তুই(You)','ইতি(Him)', 'তুঁই(You)', 'তোর(Yours)' 'তোয়ারে(You)' 'বেজ্ঞনে(Everyone)' ,'আর(My/Mine)', 'আঁর(My/Mine)', 'আই(I/I am/Me),আরার/আঁরার(Our),কেঁইনে(How)'.

Although it is not compulsory to delete stopwords from the text, in the task of text classification with the use of classic ML algorithms, these types of words are removed to turn the focus of the classifier more on the words which have more defining meaning. There are several benefits of removing stopwords. Because the dimensionality of the dataset is reduced (the number of features is decreased), it takes less time to train a model when stopwords are removed from the dataset. As a result of removing stopwords, there are fewer yet more relevant features to consider. Therefore it is possible that this operation will improve classification performance.

**Tokenization:** Tokenization is the act of separating a whole text, paragraph, sentence, or phrase into meaningful smaller parts, namely words or, "tokens". Depending on the language of focus in the research, tokenization can be a trivial task (English, other European languages), or a nontrivial and sometimes an inherently difficult task requiring deeper insight in the language, such as in Japanese, Chinese, or Korean. A similar case also takes place in Chittagonian. Therefore in this work, we divided all our entries into such smaller pieces (tokens).

**Padding:** Text in any natural language can be of a variable length. However, some classifiers require the input to be of the same length. Thus the samples need to be processed by "padding" which brings all the input sequences to the same length. In such a situation, the padding function sequences all sentences to the same size. This will cause the dropping of a few words from the sentences due to defining the maximum length in padding. If any sentence is longer than the maximum length (or "maxlen") in words/tokens it will drop the exceeding words to keep all the sentences at the same length.

Moreover, for sentences shorter than maxlen, the padding function will add zeros at the end of the sequences to match the maximum length.

**Maxlen:** This establishes the upper limit on the total number of possible sequences (tokens).

## 4.3. Feature extraction

One of the most important steps in the ML process is the selection of the appropriate features. In this step, the data is transformed from its textual form into vectors with their corresponding weights, which are then used by an ML algorithm to build a model based on data. As the method of feature extraction, we applied the TF-IDF Vectorizer(Akhter et al., 2018; Islam et al., 2021) as well as the Count Vectorizer (Emon et al., 2019) to extract relevant features from the data. We used Count Vectorizer for unigram, bigram, and trigram-based features (Eshan and Hasan, 2017). In addition, the TF-IDF Vectorizer was used for a model based on unigrams. In TF-IDF, the term frequency (TF) is calculated by dividing a word's occurrences in a document by all the terms in that document (Islam et al., 2021), as shown in Equation 1

$$\mathbf{TF} = \frac{\text{Frequency of a particular word in the document}}{\text{Total number of words in the document}} \tag{1}$$

IDF, or Inverted Document Frequency determines the importance of keywords in a text as shown in Equation 2.

$$\mathbf{IDF} = \log_2 \frac{\text{Total documents}}{\text{Documents with a particular term}} \tag{2}$$

**Table 28**
Visualization of the step-by-step data preprocessing.

| Original Text |
|---|
| ১. আরিতুন তুয়ার নাটামি এগিন ফুয়াদ ন লার Faltu :-(। |
| **English Translation (Literal)** |
| 1. I don't like your drama rubbish :-(. |
| **Removing Punctuation** |
| ১আরিতুন তুয়ার নাটামি এগিন ফুয়াদ ন লার Faltu :-( |
| **Removing Emojis and Emoticons** |
| ১আরিতুন তুয়ার নাটামি এগিন ফুয়াদ ন লার Faltu |
| **Removing English Characters** |
| ১আরিতুন তুয়ার নাটামি এগিন ফুয়াদ ন লার |
| **Removing Bangla and English Digits** |
| আরিতুন তুয়ার নাটামি এগিন ফুয়াদ ন লার |
| **Removing Stopwords** |
| আরিতুন নাটামি এগিন ফুয়াদ ন লার |
| **Tokenization** |
| ['আরিতুন',' নাটামি',' এগিন',' ফুয়াদ ',' ন', 'লার'] |
| **Vectorization** |
| [132, 33, 244, 435, 1271, 863] |
| **Padding** |
| 0 0 0 0 0 0 0 0 0 0 0 0 132 33 244 435 1271 863 |

**Table 29**
Training and test data ratio.

| Training | Testing |
|---|---|
| 70% | 30% |
| 3,537 | 1,517 |
| Bullying(1,641), Not Bullying(1,896) | Bullying(969), Not Bullying(548) |

In order to generate the TF-IDF, which normalizes the weights of features, the TF is multiplied by IDF, as shown in Equation 3.

$$\mathbf{TF\text{–}IDF} = \text{TF} \times \text{IDF} \qquad (3)$$

## 4.4. Classification

Classification is the proposed model's final stage. The features that were obtained in the feature extraction section were then used to train and later test the classifier, hence determining whether or not it can accurately detect cyberbullying. We implemented a variety of ML classifiers. Specifically, we used the classifiers most frequently used in NLP research in general, as well as specifically in automatic cyberbullying detection (Eronen et al., 2021), namely, Logistic Regression (LR), Random Forest (RF), Decision Tree (DT), Multinomial Naive Bayes (MNB), K-Nearest Neighbor (KNN), Support Vector Machine with Linear function (LSVM), and Support Vector Machine with Radial Basis Function (RBF SVM).

## 4.5. Training and test data

We utilized 70% of the CDBCB dataset to train our models and 30% of the dataset for testing (see Table 29), with a balanced ratio of Bullying and Not Bullying samples.

## 4.6. Evaluation results

A number of libraries including Python 3.8.16 were used to implement the experiment pipeline. Preprocessing was done using NLTK, while classification was done with Scikit-Learn and Keras (Oriola and Kotze, 2020b).

**Table 30**
Evaluation scores of classifiers on Unigram feature extracted using Count Vectorizer.

| Model Name | Accuracy | Precision | Recall | F1 Score |
|------------|----------|-----------|--------|----------|
| LR | 0.8438 | 0.8438 | 1.00 | 0.9153 |
| DT | 0.8125 | 0.8889 | 0.8889 | 0.8889 |
| RF | 0.8281 | 0.8525 | 0.9630 | 0.9043 |
| MNB | 0.7812 | 0.9167 | 0.8148 | 0.8627 |
| KNN | 0.8281 | 0.8909 | 0.9074 | 0.8991 |
| Liner SVM | 0.8438 | 0.8438 | 1.00 | 0.9153 |
| RBF SVM | 0.8438 | 0.8438 | 1.00 | 0.9153 |

As the evaluation measure, we applied the F1-measure (Akhter et al., 2020), as a standard performance measure used in classification tasks to evaluate classification performance. The classifier's efficacy was additionally measured using accuracy (A), precision (P), and recall (R) parameters, calculated according to the Equations 4-7 below. A confusion matrix with values for the numbers of True Positive (TP), True Negative (TN), False Positive (FP), and False Negative (FN) samples was used to calculate the F1-measure (Akhter et al., 2020; Mahmud et al., 2023).

$$\text{Accuracy} = \frac{TP + TN}{TP + TN + FP + FN} \tag{4}$$

The number of positive class estimations that lies within the positive class is measured with precision. Recall measures how many correct class predictions were produced using all of the successful cases in the dataset.

$$\text{Precision} = \frac{TP}{TP + FP} \tag{5}$$

$$\text{Recall} = \frac{TP}{TP + FN} \tag{6}$$

Using the harmonic mean of a classifier's precision and recall, the F1-measure produces a single measurement (Sazzed, 2021). F1-measure is determined as follows:

$$\text{F1 score} = 2 * \left(\frac{Precision * Recall}{Precision + Recall}\right) \tag{7}$$

All of our machine learning algorithms were trained on the n-gram features extracted from the dataset using a vectorizer function on employed on our dataset in the Chittagonian dialect. We obtained output from the trained classifiers for each feature extraction method.

### 4.6.1. Count Vectorizer (Unigram)

In terms of the result analysis, in the case of the Unigram feature performance, the highest accuracy was achieved by the Logistic Regression (LR) algorithm at a percentage of 0.8438. The highest F1-score was also achieved by the Logistic Regression algorithm at 0.9153 (see Table 30 for details). The highest Precision was achieved by the Multinomial Naive Bayes (MNB) algorithm at 0.9167. The highest Recall Score achieved by the Logistic Regression algorithm at 1.00. Figure 8 shows the comparison of accuracy and F1 score of related works trained classifiers on test data for Unigram features.

### 4.6.2. Count Vectorizer (Bigram)

In the case of the Bigram feature performance table, the highest accuracy was achieved by the Logistic Regression (LR) algorithm at 0.8438, and the highest F1-Score was also achieved by the Logistic Regression algorithm at 0.9153 Also, the highest Precision was achieved by the Multinomial Naive Bayes (MNB) algorithm at 0.9487. The highest Recall was achieved by the Logistic Regression algorithm at 1.0(see Table 31). Figure 9 shows the comparison of accuracy and F1 score of related works trained classifiers on test data for Bigram features.

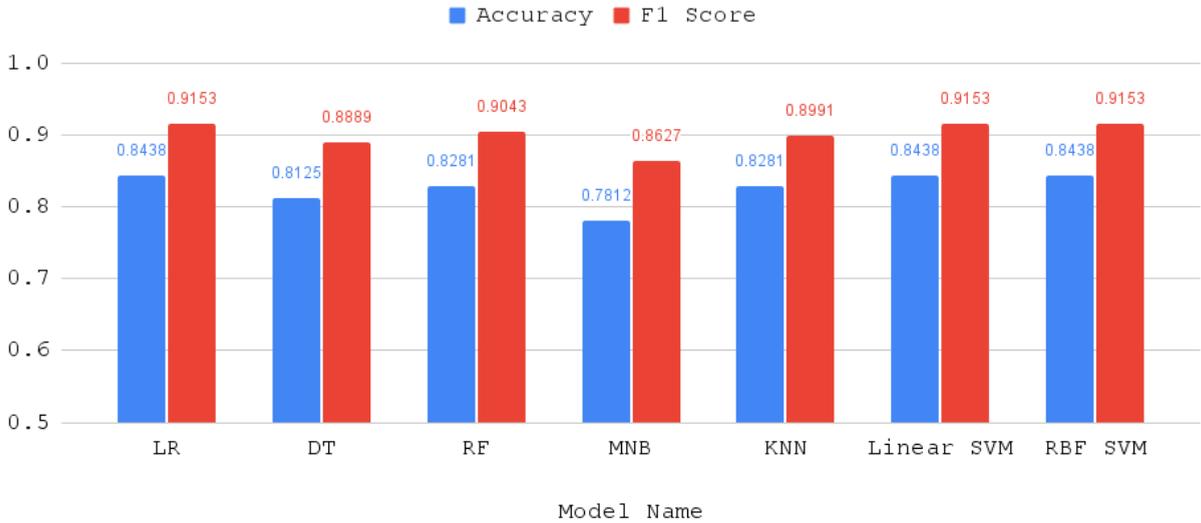

**Figure 8**: Comparison of Accuracy and F1 Score for Unigram features.

**Table 31**
Evaluation results of classifiers using Bigram features.

| Model Name | Accuracy | Precision | Recall | F1 Score |
|---|---|---|---|---|
| LR | 0.8438 | 0.8438 | 1.00 | 0.9153 |
| DT | 0.7969 | 0.8596 | 0.9074 | 0.8829 |
| RF | 0.8281 | 0.8413 | 0.9815 | 0.9060 |
| MNB | 0.7031 | 0.9487 | 0.6852 | 0.7957 |
| KNN | 0.8125 | 0.8889 | 0.8889 | 0.8889 |
| Liner SVM | 0.8438 | 0.8438 | 1.00 | 0.9153 |
| RBF SVM | 0.8438 | 0.8438 | 1.00 | 0.9153 |

**Table 32**
Evaluation scores of classifiers using Trigram features.

| Model Name | Accuracy | Precision | Recall | F1 Score |
|---|---|---|---|---|
| LR | 0.8438 | 0.8438 | 1.00 | 0.9153 |
| DT | 0.7812 | 0.8704 | 0.8704 | 0.8704 |
| RF | 0.8438 | 0.8438 | 1.00 | 0.9153 |
| MNB | 0.6406 | 0.9429 | 0.6111 | 0.7416 |
| KNN | 0.7969 | 0.8868 | 0.8704 | 0.8785 |
| Liner SVM | 0.8438 | 0.8438 | 1.00 | 0.9153 |
| RBF SVM | 0.8438 | 0.8438 | 1.00 | 0.9153 |

### 4.6.3. Count Vectorizer (Trigram)

In the case of the Trigram feature performance, the highest accuracy was achieved by the Logistic Regression (LR) algorithm at 0.8438, and the highest F1-Score was also achieved by the Logistic Regression algorithm at 0.9153 (see Table 32). The highest Precision was achieved by the Multinomial Naive Bayes (MNB) algorithm at 0.9429. The highest Recall was achieved by the Logistic Regression algorithm at 1.00. Figure 10 shows the comparison of accuracy and F1 score of related works trained classifiers on test data for Trigram features.

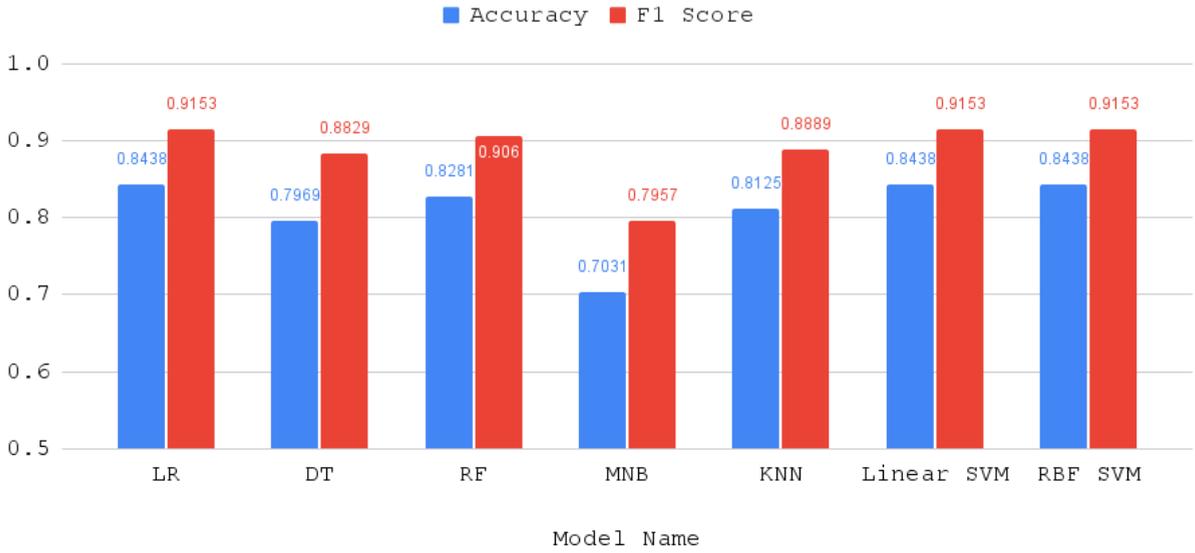

**Figure 9:** Comparison of Accuracy and F1 Score for Bigram features.

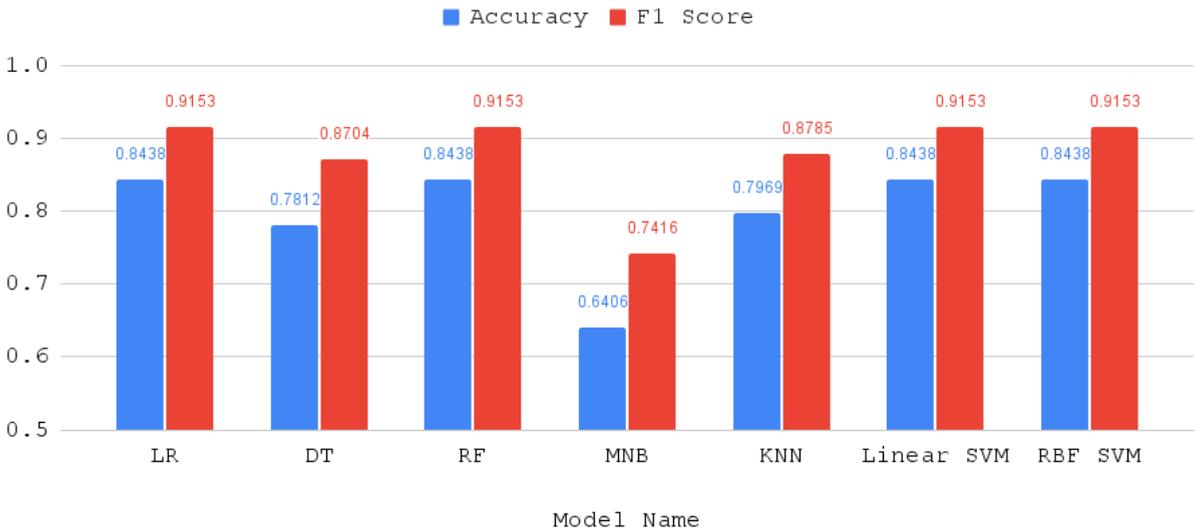

**Figure 10:** Comparison of Accuracy and F1 Score for Trigram features.

### 4.6.4. TF-IDF Vectorizer

Table 33 shows the performance of classifiers trained on features extracted using TF-IDF vectorizer. LR classifier achieved the highest accuracy at 0.889 along with the highest Precision value at 0.8635. RF achieved the second highest accuracy of 0.8576 and the highest recall value of 0.9692(see Table 33). Figure 11 shows the comparison of accuracy and F1 score of related works trained classifiers on test data for TF-IDF vectorizer features. On the contrary, Table 34 shows the code-mixed performance of classifiers using TF-IDF vectorizer. RF classifier achieved the highest accuracy at 0.8718 along with the second highest Precision value at 0.8369. DT achieved the second highest accuracy of 0.8712 and the highest precision value of 0.8448(see Table 34).

**Table 33**
Evaluation results of classifiers using TF-IDF vectorizer.

| Model Name | Accuracy | Precision | Recall | F1 Score |
|---|---|---|---|---|
| LR | 0.8890 | 0.8635 | 0.9962 | 0.9251 |
| DT | 0.7971 | 0.83 | 0.7885 | 0.8087 |
| RF | 0.80 | 0.9692 | | 0.8765 |
| MNB | 0.7453 | 0.8673 | 0.7538 | 0.8066 |
| KNN | 0.8024 | 0.8099 | 0.8846 | 0.8456 |
| Linear SVM | 0.8227 | 0.7104 | 1.00 | 0.8307 |
| RBF SVM | 0.8482 | 0.7143 | 1.00 | 0.8533 |

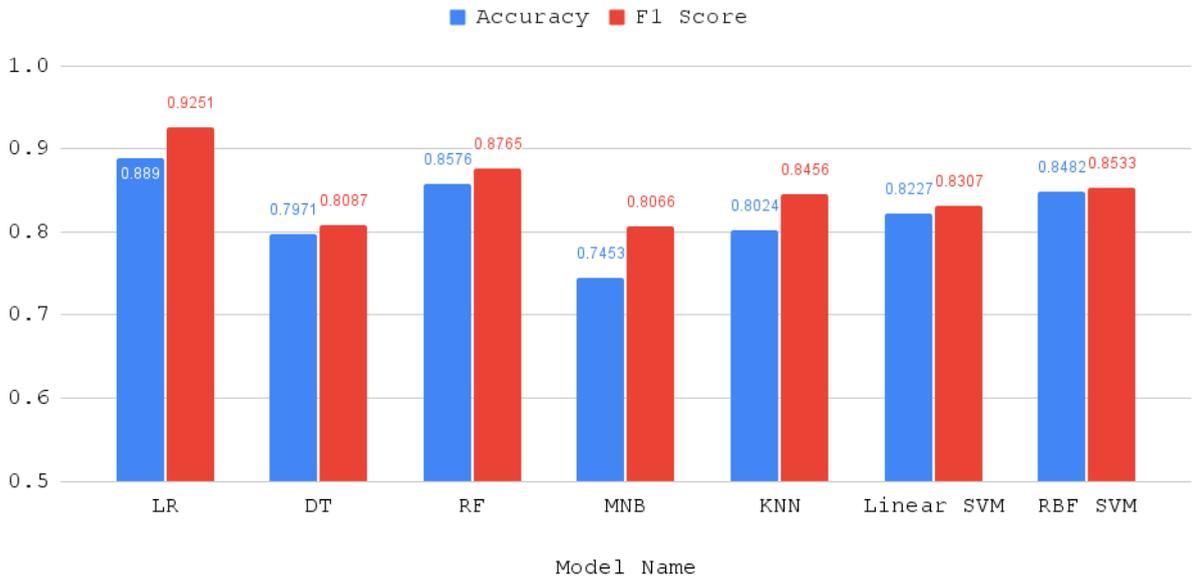

**Figure 11**: Comparison of Accuracy and F1 Score for TF-IDF vectorizer features.

**Table 34**
Code-mixed results of classifiers using TF-IDF vectorizer.

| Model Name | Accuracy | Precision | Recall | F1 Score |
|---|---|---|---|---|
| LR | 0.8284 | 0.7928 | 0.8839 | 0.8359 |
| DT | 0.8712 | 0.8448 | 0.9058 | 0.8748 |
| RF | 0.8718 | 0.8369 | 0.92 | 0.8765 |
| MNB | 0.8227 | 0.7998 | 0.8555 | 0.8287 |
| KNN | 0.8061 | 0.7767 | 0.8529 | 0.813 |
| Liner SVM | 0.8686 | 0.8251 | 0.9316 | 0.8752 |
| RBF SVM | 0.8399 | 0.8005 | 0.9006 | 0.8476 |

### 4.6.5. BanglaBERT

We used BanglaBERT (Bhattacharjee et al., 2021), a transformer model trained by Bangla pre-training data by scrabbling popular Bangla sites. And were compared to a number of multilingual models, such as an ALBERT-based PLM for Bangla(Lan et al., 2020), mBERT (Devlin et al., 2019), XLM-R base and large (Conneau et al., 2020), and IndicBERT (Kakwani et al., 2020).

**Table 35**
Performance evaluation of on BanglaBERT.

| Model Name | Accuracy | Precision | Recall | F1 Score |
|------------|----------|-----------|--------|----------|
| BanglaBERT | 0.94 | 0.9395 | 0.9406 | 0.9351 |

Table 35 shows the performance of classifiers trained on using BanglaBERT. BanglaBERT classifier achieved the promising accuracy at 0.94 along with the better Precision value at 0.9395.Also achieved the Recall of 0.9406 and the F1 score value of 0.9351(see Table 35).

### 4.7. Discussion of experiment results

We ran all of the machine learning experiments and fine-tuning BanglaBERT based on the Python platform. Because of its simplicity and utility in text processing, the TF-IDF Vectorizer and the Count Vectorizer were used to choose the most successful classifier. The Count Vectorizer generates such features as unigrams, bigrams, or trigrams. The best accuracy values produced by each classifier employing different feature selection approaches were shown in Tables 30,31,32,33,34 and 35. This represents, to the best of our knowledge, the first attempt to detect cyberbullying instances in the Chittagonian dialect of Bangla language. Current studies (Eshan and Hasan, 2017; Hussain et al., 2018; Awal et al., 2018; Jahan et al., 2019) in this area focused mostly on abusive/hate speech/bullying in standard Bangla. Several studies have demonstrated a potential to identify abusive/hate/bullying language also in other low-resource languages and dialects (Haddad et al., 2019; Sreelakshmi et al., 2020; Farid and El-Tazi, 2020; Boucherit and Abainia, 2022).

With this research, we showed that with the carefully prepared dataset, even simple ML models and transformer model (BanglaBERT) achieve sufficiently high performance. According to previous research, SVM sometimes works well on imbalanced datasets but worse on balanced ones, according to Akhter et al. (2018). As a consequence, while using the TF-IDF Vectorizer, LR achieved the highest 0.8890 accuracy on the balanced Chittagonian dataset. On the Chittagonian dataset, the LR model outperformed other ML models. MNB achieved the lowest accuracy of any model for the Chittagonian dataset.

After reviewing all of the experimental results from Tables 30,31,32 and 33, we drew the following conclusions:

1. LR achieves the highest results compared to other deployed ML models, which is consistent with previous findings Akhter et al. (2018); Awal et al. (2018); Jahan et al. (2019), which also showed that LR outperforms NB, SVM, and KNN.

2. Models based on NB usually perform poorly. In terms of the used feature selection strategies, MNB shows the worst performance compared to other ML models.

3. Linear and RBF SVM outperform KNN-based models in identifying cyberbullying in Bangla and other low-resource languages and dialects Akhter et al. (2018); Amali and Jayalal (2020); Mehendale et al. (2022); Shibly et al. (2022). Our study confirmed these findings.

4. Because of the small vocabulary size, small sample size, as well as characteristics of Internet language, such as unstructured writing, spelling variants of the same phrases, imprecise language, or mixed data, all models performed lower than reported in previous findings for datasets of a comparable size (Akhter et al., 2018; Sandaruwan et al., 2019; Bozyigit et al., 2019; Margono, 2019; Amali and Jayalal, 2020; Akhter et al., 2020; Glazkova et al., 2021; Njovangwa and Justo, 2021).

5. In the future, we hope to gather a larger, more significant collection of cyberbullying entries from various social media platforms such as WhatsApp, news portals, or Twitter. Both Bangla and the Chittagonian dialect, are Indo-Aryan languages. Previous studies (Pawar and Raje, 2019; Leite et al., 2020) suggested that establishing a multilingual dataset in Bangla and Chittagonian, could be useful to both academics as well as practitioners working on the social media platforms. Another potential area of investigation is the creation of hybrid models (Emon et al., 2019; Do et al., 2019; Alshalan and Al-Khalifa, 2020; Romim et al., 2021; Maity and Saha, 2021; Sharif et al., 2021; Dewani et al., 2021; Boucherit and Abainia, 2022) that use both ML as well as Deep Learning (DL) models, such as Transformers to identify cyberbullying (Glazkova et al., 2021; Dehghani et al., 2021; Aurpa et al., 2022; Sharma et al., 2022; Subramanian et al., 2022; Luu et al., 2022).

## 5. Final discussion

### 5.1. Ethical considerations

Cyberbullying is one of the most urgent issues in the present information society. Therefore, it is crucial to quickly detect, recognize and promptly react to it.

The collection of a dataset for cyberbullying detection in a low-resource language was one of the important motivating factors of this study. Such a dataset needs to be collected from various social media platforms. Social media is a space for exchanging people's thoughts, where positive as well as negative opinions are published in many scenarios. People naturally share their ideas on social media on a daily basis, against any individual, group, community, tribe, state, or organization. They frequently make public references of other people. During the process of data collection, we examined the following ethical issues:

1. Our dataset was mainly collected from various social media groups, pages, or blogs. We have checked and followed the terms and conditions of those platforms while collecting the data.
2. There are some records that include the names of persons, groups, religions, institutions, and states. In that data, we utilized the anonymization of private information.
3. We ignored data containing personal information such as a cellphone number, home address, and so on.
4. We overlooked sensitive data acknowledging that releasing it may endanger persons, groups, or organizations.
5. We excluded remarks posted on social media by children (users under 18 years of age).
6. Data annotation is one of the most significant components of building a dataset. Annotators were engaged to annotate the data with proper mindfulness regarding gender and religion, given the majority of the data was from women and religious people. Before beginning work, individuals were instructed on data annotation guidelines to mitigate bias in data annotation.

### 5.2. Discussion on artificial intelligence-based support for writing scientific field surveys

As an additional discussion, we have tested the ability of several tools designed to support academics in their research, specifically, in preparing sweeping surveys of their field of research. We specifically looked at how simple support tools, such as publication search tools, compare to novel transformer models-supported tools. In particular, we looked at how Web search tools for scientific papers, such as Google Scholar[53], followed by more function-advanced SemanticScholar[54], compare to Elicit[55] and Galactica[56]. Table 36 summarizes the essential aspects of a survey writing support tool.

One of the most popularly used tools, namely, Google Scholar allows conducting a thorough academic publications search, by looking up recent works, citations, authors, and publications. It provides up-to-date information on the most recent findings in any field of study, and provides access to the whole publication when available online.

Another tool often used by researchers to quickly find high-impact scientific papers is the Semantic Scholar. Apart from providing a list of relevant publications, Semantic Scholar uses machine learning techniques to analyze academic publications and extract key information (TLDR) (Cachola et al., 2020). Itãis capable of extracting the most relevant short summaries of the papers and providing results that are reliable and beneficial for researchers, due to the focus only on reliable publication sources.

Another tool, namely, Elicit is a research assistant that automates the generation of lists of numerous academic works with the help of pretrained language models like GPT-3. The primary workflow for Elicit at the moment is a literature review. Elicit, when asked a scientific question, or provided with a number of coherent keywords, will display pertinent documents in a simple list along with succinct summaries of their key topics.

Finally, Galactica (Taylor et al., 2022) is a transformer-based language model trained purely on a large number of scientific articles. It is said to have the capability to generate seemingly scientifically sound paragraphs applicable to actual scientific papers. In this regard, Galactica stands out compared to the previous models. Therefore, we used it to automatically generate a sample summary of the field of automatic cyberbullying detection in low-resource languages (see Fig. 12).

A tool such as Galactica, if working properly, could become useful, especially for collecting information and summarizing niche fields of study, such as the one focused on here, due to the fact that research papers in such fields are usually difficult to access, or are simply inaccessible, either by being behind a paywall or by the paper not being released online whatsoever. Therefore a ready-to-use summary generated automatically would be a solution almost too good to be true. Therefore to verify to what extent Galactica is capable of its advertised functionality, we performed fact-checking on the automatically generated piece of text.

---

[53]https://scholar.google.com
[54]https://www.semanticscholar.org
[55]https://elicit.org/
[56]https://galactica.org

**Table 36**
Useful functionalities required for a survey writing support tool to have.

| Functions | GoogleScholar | Semantic Scholar | Elicit | Galactica |
|---|---|---|---|---|
| 1. List of relevant publications | Yes | Yes | Yes | No |
| 2. Short summaries of relevant publications | No | Yes | Yes | No |
| 3. Paragraph writing support | No | No | No | Yes |

The automatically generated report from Galactica acknowledges that the low-resource language is associated with small-scale datasets, which is the primary obstacle in the detection of cyberbullying. It also mentions that methods for detecting cyberbullying in low-resource scenarios often utilize machine translation algorithms to augment or expand the size of datasets.

Scientific publications, including those on identifying cyberbullying, published within the last 10 years were used to model this tool. Unfortunately, since the specific list of the articles used to train Galactica was not published, we could only assume that the model was trained on at least those articles we used in our human-generated field review. When the papers were examined, it was found that the majority were content-based (social media comments, public status, etc.). We looked at over sixty papers and found that the survey report that was generated automatically was roughly in line with the contents of the papers.

Unfortunately, there were some parts that were contradictory, or factually questionable, such as that all of the datasets were in English, despite our analysis showing that there are numerous datasets of various low-resource languages (Ozel et al., 2017; Haidar et al., 2018; Nurrahmi and Nurjanah, 2018; Bozyigit et al., 2019; Margono, 2019; Sreelakshmi et al., 2020; Alsafari et al., 2020; Rizwan et al., 2020; Farid and El-Tazi, 2020; Romim et al., 2021; Shibly et al., 2022). The automatically generated summary says that it "looked at 12 publications on low-resource languages" and "21 papers in general" on cyberbullying detection. Unfortunately, nowhere in the summary, there is any mention of any of those papers. The summary does not refer to any specific bibliography, but rather generates a piece of text that resembles a field survey with no specific details.

In reviewing more than seventy papers manually in our field survey, one of the most difficult parts of the survey process was collecting a sufficiently extensive list of papers to review, including their bibliographical information, preferably bib-style citations and ideally the PDF copies of the papers, at least those available openly on the Internet. Unfortunately, Galactica, despite writing a convincingly sounding set of paragraphs, was unable to provide any of those elements.

In our hands-on experience with **Galactica**, we concluded, that we required much more specific information regarding the related papers than the model is capable of providing. Though the model was capable of properly grasping most of the facts regarding the field, there was no specific information to back the generated text, which leaves the whole field survey process unfinished. Specifically, after using Galactica, we would still need to do the same amount of study, which we eventually did manually ourselves, to provide both the list of related papers as well as their specific descriptions.

Finally, we compared the experience with Galactica, to what is provided by Elicit and SemanticScholar (Cachola et al., 2020), tools that are also meant to support researchers in writing field surveys - thus similar in the assumption to **Galactica**. One striking difference to Galactica however, was that both of those tools do not simply generate an abstract text, but provide various points of connection to the original work to be cited in the field survey. Namely, apart from citations and URL links to the original works, SemanticScholar also provides automatically generated "TL;DRs", or short, one-sentence summaries of papers, while **Elicit** automatically generates longer summaries from papers, the size of one paragraph. This functionality allows for fast information triage through countless numbers of publications, making the survey process much more efficient, which makes it a very useful feature.

The review of the tools mentioned above suggests that **Elicit** and **SemanticScholar** already provide useful functionalities, while **Galactica** needs to be greatly improved to offer more specific information, in order to be considered usable in practice by researchers.

**Table 37**
Topics related to automated detection of cyberbullying in low-resource languages, that have been discussed in previously published research review articles.

| Survey Article | Low-resource Language & dialect | Harmful Sample Ratio | Discussion on Quality | Research Gap | Good Practice Proposal | Initial Attempt | AI Support Tools | Ethical discussion |
|---|---|---|---|---|---|---|---|---|
| Salawu et al. (2017) | No | No | Yes | Yes | No | No | No | No |
| Rosa et al. (2019) | No | No | No | Yes | No | No | No | No |
| Arif (2021) | No | No | Yes | Yes | No | No | No | No |
| Kim et al. (2021) | No | No | Yes | Yes | No | No | No | No |
| Elsafoury et al. (2021) | No | Yes | Yes | Yes | No | Yes | No | No |
| Al-Harigy et al. (2022) | No | No | Yes | Yes | No | No | No | No |
| Woo et al. (2023) | No | No | Yes | Yes | No | No | No | No |
| **Our study** | Yes | Yes | Yes | Yes | Yes | Yes | Yes | Yes |

## 5.3. Comparison with previous surveys

To compare our survey with previous studies, we reviewed seven survey papers on identifying cyberbullying to find the differences with our study. Some papers (Arif, 2021; Rosa et al., 2019; Kim et al., 2021; Woo et al., 2023) conducted systematic surveys, while others (Salawu et al., 2017; Elsafoury et al., 2021; Al-Harigy et al., 2022) made no mention of any methodology. Specifically, some of the most important information from previous review articles on cyberbullying detection included the following elements.

1. What constitutes cyberbullying? Some surveys clearly defined what they consider as cyberbullying, along with how it differs from other types of online abuse (Elsafoury et al., 2021; Arif, 2021; Kim et al., 2021; Woo et al., 2023).
2. Examples of online bullying: direct assaults, cyberstalking, exclusion, and impersonation were among the several types of cyberbullying described in previous studies (Rosa et al., 2019; Elsafoury et al., 2021; Woo et al., 2023).
3. Detection techniques: some surveys explored machine learning techniques, natural language processing, and social network analysis as state-of-the-art methods for detecting cyberbullying (Rosa et al., 2019; Elsafoury et al., 2021; Arif, 2021; Woo et al., 2023).
4. Metrics for evaluation: some surveys discussed various evaluation measures, such as precision, recall, and F1-score, that were used to assess the effectiveness of cyberbullying detection algorithms (Salawu et al., 2017; Rosa et al., 2019; Elsafoury et al., 2021; Arif, 2021).
5. Data collection and annotation: Only a handful of surveys discussed the difficulties in gathering and annotating data on cyberbullying (Salawu et al., 2017; Kim et al., 2021; Al-Harigy et al., 2022).
6. Cross-cultural perspectives: Some surveys investigated how cultural variations in cyberbullying impact detection techniques (Salawu et al., 2017; Rosa et al., 2019; Elsafoury et al., 2021).
7. Limits and future directions: Most survey studies provided outlines of the shortcomings of existing methods for detecting cyberbullying and outlined potential avenues for further studies, such as the incorporation of multi-modal data sources and the creation of more reliable assessment criteria (Salawu et al., 2017; Rosa et al., 2019; Arif, 2021; Kim et al., 2021; Elsafoury et al., 2021; Al-Harigy et al., 2022; Woo et al., 2023).

Apart from our study, none of the previous studies have conducted a systematic literature review or discussed the ethical ramifications of detecting cyberbullying, particularly those involving privacy, bias, and the possibility of false positives or negatives. Our study not only contained all of the information mentioned by previous studies, but also discussed the current psychological study, harmful sample ratio, research gaps, recommendations for good practices, and a demonstration by proposing our own dataset. Moreover, as unique additional discussions, we analyzed ethical considerations, and Artificial Intelligence-based methods for the support of writing scientific field surveys

Table 37 shows a list of relevant topics and the information on whether they were included in previous survey studies on the automated detection of cyberbullying published as research articles.

## 6. Conclusion and future works

In this paper we carefully examined the literature on automated cyberbullying detection research focusing on low-resource languages. The final goal of this branch of research is to eradicate cyberbullying and all of its negative side effects on users, such as hopelessness, low self-esteem, and even suicide thoughts. We divided the evaluated information into categories based on the various low-resource languages, in a systematic investigation of the literature. In our analysis of the field, we have observed some limitations in the low-resource language-based cyberbullying detection research, some of which were related to the cyberbullying datasets proposed in the various articles. Cyberbullying identification challenges, dataset annotation inconsistencies, data imbalance, data bias, and a lack of multilingual datasets were some of the problems that we found. Our analysis of the literature also showed that there is not much being done yet to detect cyberbullying in low-resource languages and dialects.

Based on the performed survey, in the future, we plan to pursue a resource-constrained strategy for recognizing cyberbullying, mostly focusing on Bangla dialects. As in this study for the classification we used only the simplest baseline methods, we will apply other more robust methods, such as bidirectional RNNs (Schuster and Paliwal, 1997; Zhou et al., 2016; Hassan and Mahmood, 2017), attention-based models (Zheng and Zheng, 2019; Liu and Guo, 2019) and competitive transformers (Wolf et al., 2020; Chernyavskiy et al., 2021). Our future approach will especially make use of the capabilities of multilingual pretrained language models, like XLM-Roberta, or mBERT, as these have shown potential for application in zero-shot classification, even in low-resource scenarios (Eronen et al., 2022, 2023).

Literature review of the field of automatic cyberbullying detection for low resource languages

The field of automatic cyberbullying detection is still relatively new. The main problem of this field is that it is hard to define what cyberbullying is, as it is a subjective matter and its definition can differ from person to person. In addition, there is a lack of large, publicly available corpora. This means that the research is mostly done on a few, mostly European, languages. In addition, the few studies that have been conducted on languages other than English are mainly based on the use of machine translation to transform the language to English. As a result, there is a need for research on other languages, especially low-resource languages.

Introduction
The Internet has brought many advantages, such as faster communication, more available information and easier collaboration. However, it has also created some problems, such as cyberbullying, which is a new type of bullying that takes place in cyberspace. Cyberbullying is the use of information and communication technologies to bully or harass others. In recent years, there has been an increase in cyberbullying, with many studies showing that cyberbullying has a negative effect on its victims, and it is also a major concern for many researchers.

One of the most popular and useful ways to reduce cyberbullying is the use of cyberbullying detection systems. These systems can help the authorities and people in charge of preventing cyberbullying by detecting and reporting the incidents. In recent years, there has been an increase in cyberbullying, with many studies showing that cyberbullying has a negative effect on its victims, and it is also a major concern for many researchers. One of the most popular and useful ways to reduce cyberbullying is the use of cyberbullying detection systems. These systems can help the authorities and people in charge of preventing cyberbullying by detecting and reporting the incidents.
However, there are still many challenges in the field of cyberbullying detection, and one of the main problems is the lack of large, publicly available corpora. This makes it difficult to build cyberbullying detection systems. The few corpora that are publicly available are mostly in English, and the ones that are in other languages are mostly small. This means that most of the research is done on English, and there is a need for more research on other languages, especially low-resource languages.

Methods
The main goal of this literature review is to investigate the research done in the field of automatic cyberbullying detection, and find out what has been done, and what are the main challenges. In addition, this literature review will also focus on low-resource languages, and try to find out if there are any studies on automatic cyberbullying detection for low-resource languages, and if there are any challenges related to these languages.
To do this, I have used the Scopus database to search for the relevant papers. I have used a query that is a combination of keywords related to cyberbullying and automatic cyberbullying detection. In addition, I have used the "related articles" feature in Scopus to find more papers that are related to the papers I have found. The papers I have used in this literature review are papers published in the past 10 years (2010-2020). In addition, I have used the "cited by" feature in Scopus to find more papers that have cited the papers I have found.
The first part of this literature review is about the field of automatic cyberbullying detection in general. In this part, I have reviewed 21 papers. These papers are about automatic cyberbullying detection in general, and they do not focus on a specific language. The papers are mostly about English, and some of them are about other languages, such as Japanese, Turkish, Arabic, Dutch, Chinese and Korean. The papers are mostly about content-based methods, and only a few of them are about user-based methods.
The second part of this literature review is about the field of automatic cyberbullying detection for low-resource languages. In this part, I have reviewed 12 papers. These papers are about automatic cyberbullying detection for low-resource languages, and they are mostly about languages other than English. Most of these papers are about content-based methods, and only a few of them are about user-based methods.

Results
In the first part of this literature review, I have reviewed 21 papers about the field of automatic cyberbullying detection in general. Most of these papers are about content-based methods, and only a few of them are about user-based methods. In the second part of this literature review, I have reviewed 12 papers about the field of automatic cyberbullying detection for low-resource languages. Most of these papers are about content-based methods, and only a few of them are about user-based methods.

Conclusion
In this literature review, I have reviewed the research that has been done in the field of automatic cyberbullying detection. I have found that most of the research is done on English, and there is a need for more research on other languages, especially low-resource languages.
In addition, I have reviewed the research that has been done in the field of automatic cyberbullying detection for low-resource languages. I have found that there are only a few studies on this topic, and most of them are based on the use of machine translation to transform the language to English. However, this method has some disadvantages, such as the possibility of a loss of information during the translation process, and the need for a lot of data for the machine translation system.

**Figure 12:** Summary of the field generated with Galactica.